\documentclass{article} 
\usepackage{iclr2027_conference,times}

\usepackage{amsmath,amsfonts,bm}

\def\eqref#1{equation~\ref{#1}}

\def\1{\bm{1}}

\DeclareMathAlphabet{\mathsfit}{\encodingdefault}{\sfdefault}{m}{sl}
\SetMathAlphabet{\mathsfit}{bold}{\encodingdefault}{\sfdefault}{bx}{n}

\usepackage{hyperref}
\usepackage{url}
\usepackage{amsmath,amsthm}
\usepackage{booktabs}
\usepackage{graphicx}
\usepackage{amssymb}
\usepackage{algorithm}
\usepackage{algpseudocode}
\usepackage{wrapfig}
\usepackage[table]{xcolor}
\usepackage{tikz-cd}
\usepackage{tikz} 
\usetikzlibrary{arrows.meta,positioning}
\usepackage{mathtools}
\usepackage[most]{tcolorbox}

\newtcolorbox{artifactbox}[1]{
enhanced,
breakable,
colback=gray!3,
colframe=black!35,
boxrule=0.5pt,
arc=1.5pt,
left=5pt,
right=5pt,
top=4pt,
bottom=4pt,
title={#1},
fonttitle=\bfseries\small,
before skip=6pt,
after skip=6pt
}

\usepackage[colorinlistoftodos]{todonotes}
\newcommand{\ms}[2]{$#1$\,{\scriptsize$\pm #2$}}

\newcommand{\cmark}{$\checkmark$}
\newcommand{\xmark}{$\times$}
\newcommand{\pmark}{$\sim$}

\title{%
\begin{minipage}[c]{0.18\textwidth}
  \includegraphics[width=\linewidth]{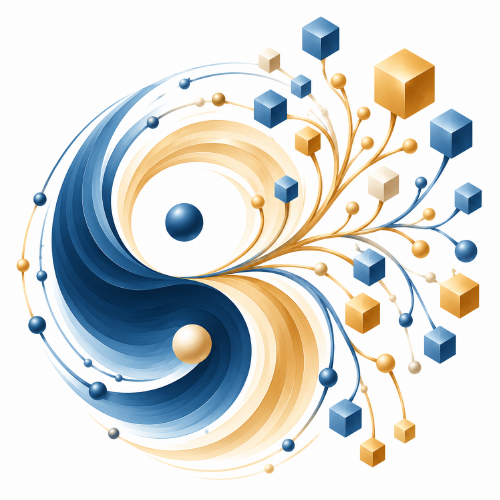}
\end{minipage}%
\hfill
\begin{minipage}[c]{0.80\textwidth}
GRADUS: Discovering and Recovering Agent Lessons from the Environment's Latent Syllabus
\end{minipage}}

\author{
Qing Ping\textsuperscript{1},
  Wei Yang\textsuperscript{2}\thanks{Work was done during an internship at AWS AI.},
  Panpan Xu\textsuperscript{1},
  Johannes Kirmayr\textsuperscript{3},
  Lukas Stappen\textsuperscript{3}, \\
  \textbf{Junjie Tang\textsuperscript{2},
  Jun Huan\textsuperscript{2}} \\
  \\
  \textsuperscript{1}AWS AI \\
  \textsuperscript{2}University of Southern California \\
  \textsuperscript{3}BMW Group Research and Technology
}

\iclrfinalcopy 
\begin{document}

\maketitle
\lhead{Preprint}

\begin{abstract}

Training LLM agents for policy-constrained tool environments, such as an airline backend with booking, refund, and cancellation policies, requires tasks that exercise the environment's rules individually and in combination. Existing synthesis pipelines let a model propose tasks and retain only those it immediately solves, biasing training data toward easier behaviors. We introduce \textsc{Gradus} (\textbf{G}raded \textbf{R}ungs of \textbf{A}scending \textbf{D}ifficulty from an \textbf{U}nderlying \textbf{S}ystem), which decouples what a task should teach from the concrete task and solution used to teach it. Without seed tasks, \textsc{Gradus} treats a fixed environment as a latent syllabus, extracting guarded behavioral operators that capture how actions depend on current conditions and how their effects reshape subsequent decisions. It composes these operators into increasingly demanding tasks while using corpus-level feedback to target missing outcome  combinations. Crucially, \textsc{Gradus} jointly calibrates tasks and solutions against the environment: execution feedback identifies which side to refine while preserving the intended behavioral requirements. Verification thus becomes a recovery signal rather than a terminal filter, coupling coverage-guided exploration with the preservation of difficult examples. From off-the-shelf backbones, the resulting curriculum lifts a 30B model from 59\% to 89\% on $\tau^2$-Bench and from 47\% to 83\% on CAR-Bench with an Opus 4.5 proposer, and to 76\% and 62\%, respectively, with an open-weight Qwen3-235B proposer. These results show the value of jointly controlling which behaviors are explored and which survive into verified training data.

\end{abstract}

\section{Introduction}

Training LLMs to act as agents requires data spanning diverse tool use, reasoning demands, and multi-turn interactions. Recent synthesis pipelines build generalist agents by generating tasks, workflows, and even entire environments at scale~\citep{lyu-etal-2026-agenticqwen,wang2025klearagentforgeforgingagenticintelligence,song-etal-2026-envscaler}. These efforts broaden the range of situations agents encounter. A complementary problem arises when the  environment is already given: how can we construct a curriculum that prepares an agent for the full range of behaviors required by that particular environment?

A fixed environment contains substantially more training structure than its list of APIs suggests. Depending on the current state and operating policies, the same action may be permitted, denied, or require additional steps. For example, in an airline booking environment, whether a cancellation is permitted depends on the cabin class, the time since booking, and whether travel insurance covers the stated reason, so the same request may end up with different agent behaviors in different states. Moreover, an earlier action may change the conditions governing later decisions. Constructing training data from this environment structure presents two coupled challenges: deciding which behaviors to teach and producing valid tasks and solutions that teach them.

A high-coverage curriculum must therefore identify these behavioral distinctions, compose them systematically, and track which remain underrepresented. Existing methods structure synthesis through task blueprints and workflows~\citep{prabhakar2025apigenmtagenticpipelinemultiturn}, explicit constraints~\citep{chen2026covetraininginteractivetooluse}, domain-entity graphs~\citep{sun2026agentskillerscalinggeneralistagent}, or predefined behavioral and difficulty dimensions~\citep{gu2026writwritereadintensivetrajectory}. These structures effectively guide individual proposals, but individually diverse or difficult tasks do not necessarily produce broad corpus-level coverage of policy outcomes and their interactions.

A second challenge is preserving the intended coverage during verification. Existing synthesis pipelines commonly use successful execution, constraint satisfaction, or exact state agreement as an acceptance gate, retaining verified trajectories while filtering unsuccessful candidates~\citep{sun2026agentskillerscalinggeneralistagent,chen2026covetraininginteractivetooluse,lyu-etal-2026-agenticqwen,zeng2026logigenlogicdrivengenerationverifiable}. Although this improves data precision, it also makes verification a hidden curriculum selector. As tasks combine more rules and state changes, errors in both task construction and solution generation become more likely, causing difficult behaviors to be disproportionately dropped. 

We introduce \textsc{Gradus} (\textbf{G}raded \textbf{R}ungs of \textbf{A}scending \textbf{D}ifficulty from an \textbf{U}nderlying \textbf{S}ystem), which separates what agent behaviors to teach and producing valid tasks and solutions that teach them. Starting from a fixed executable environment without seed tasks, \textsc{Gradus} organizes each action according to the conditions governing its use and the consequences of its possible outcomes. These reusable descriptions, which we call \emph{guarded operators}, expose how one action can enable, invalidate, or otherwise alter the requirements of another. \textsc{Gradus} composes these relationships into increasingly demanding task blueprints, while corpus-level feedback directs generation toward underrepresented outcomes and combinations. Additional cross-space interaction dimensions vary how the same underlying requirements are surfaced by introducing ambiguity, indirect references, social pressure, or reasoning demands.

Crucially, as blueprints become more difficult, some fail to yield a valid task and solution on the first attempt, and neither the proposed task nor its solution is assumed correct. Using the executable environment as a common anchor, \textsc{Gradus} lets task design and solution construction calibrate each other: failures in solution construction feed back into task revision, flagging unsolvable tasks, hallucinated states or conditions, and ambiguous objectives, while the task and its oracle proposal anchor the solution search, steering it to fix departures from policy. Both sides are refined against environment execution, helping difficult compositions  survive verification rather than being discarded.

From off-the-shelf backbones, the resulting curriculum lifts a 30B model from $59\%$ to $89\%$ on $\tau^2$-Bench and from $47\%$ to $83\%$ on CAR-Bench with an Opus~4.5 proposer, and to $76\%$ and $62\%$, respectively, with an open-weight Qwen3-235B proposer. These results demonstrate the value of jointly controlling which behaviors are explored and which survive into verified training data. Our contributions are:
\begin{itemize}
    \item A compositional representation of environment behavior that connects action conditions, policy-dependent outcomes, and evolving state.
    \item Coverage-guided task construction with explicit controls over structural complexity and interaction demands.
    \item Environment-grounded joint task-solution refinement that recovers difficult examples while preserving their intended behavioral requirements, connecting proposed coverage to verified training data.
\end{itemize}

\section{Methodology}
\label{sec:method}

\begin{figure*}[t]
      \centering
      \includegraphics[width=\textwidth]{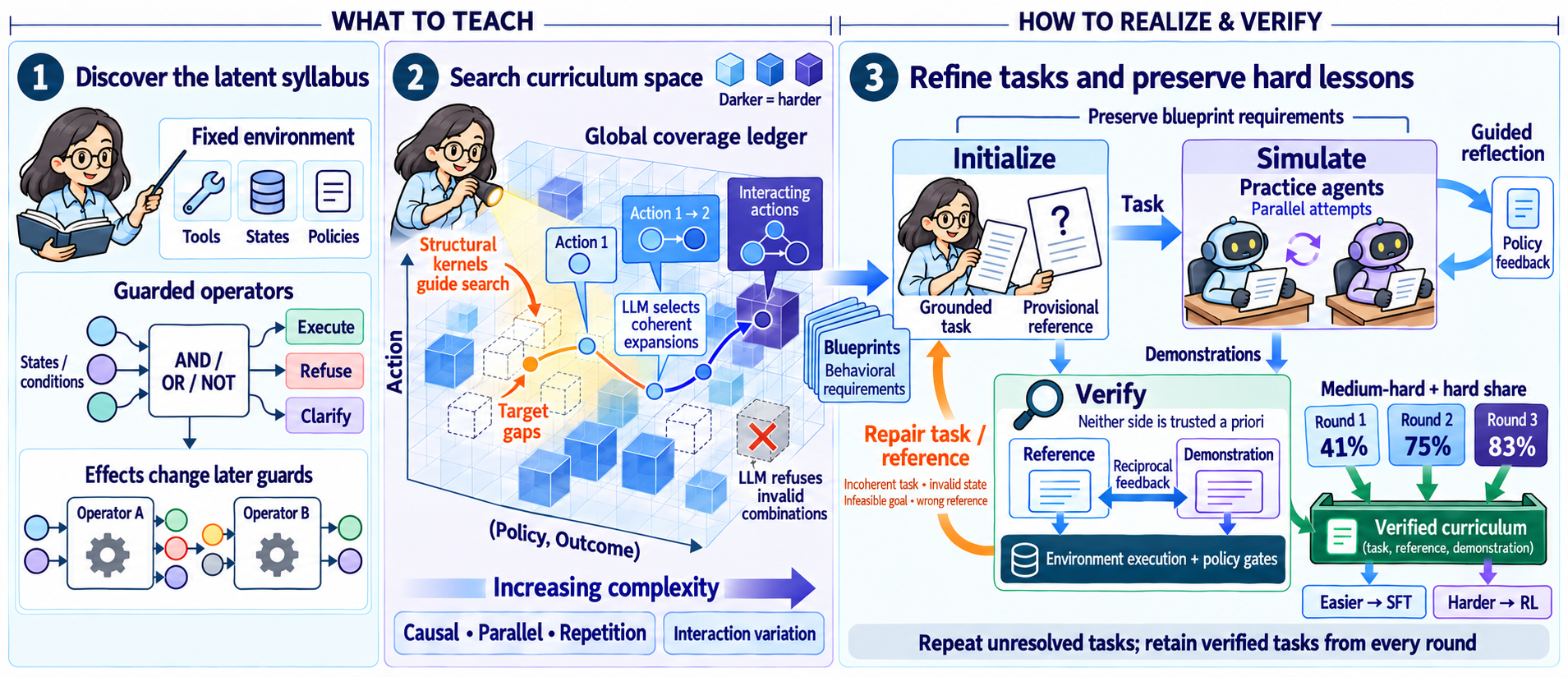}
      \caption{\textbf{Overview of \textsc{Gradus}.}
      \textbf{(1)} Environment artifacts are compiled into guarded operators.
      \textbf{(2)} Coverage-guided evolution searches uncovered syllabus regions.
      \textbf{(3)} Joint task-solution refinement recovers unresolved candidates,
      preserving difficult lessons while accumulating verified examples.}
      \label{fig:gradus-overview}
  \end{figure*}

\subsection{The Latent Syllabus}

A fixed environment implicitly specifies what an agent must learn through its tools, states, policies, and executable transitions. In an airline system, for example, the same cancellation request may require a refund or a refusal depending on the booking state and policy; earlier booking changes may alter which options remain available. These conditional behaviors and their interactions form the environment's \emph{latent syllabus}. They are distributed across system artifacts rather than supplied as an explicit collection of training tasks.

\textsc{Gradus} turns this implicit syllabus into verified training data through

\begin{equation}
  \label{eq:GRADUS-overview}
  E
  \xrightarrow{\text{\scriptsize Organize}}
  \mathcal G
  \xrightarrow{\text{\scriptsize Coverage-guided search}}
  \mathcal Z
  \xrightleftharpoons[
    \text{\scriptsize Diagnose and re-design}
  ]{
    \text{\scriptsize Construct and verify}
  }
  \mathcal C
  \xrightarrow{\text{\scriptsize Accept}}
  \mathcal D .
  \end{equation}

where $G$ contains reusable descriptions of environment behaviors, $Z$ contains task blueprints, $C$ contains the realized curriculum, and $D$ is the resulting accepted training corpus. A \emph{guarded operator} in $G$ describes an action together with the conditions governing its possible outcomes, such as when cancellation is allowed and what refund follows. \emph{Coverage-guided search} composes these descriptions into blueprints that specify which behaviors a task should exercise, prioritizing underrepresented outcomes and interactions. A blueprint might combine a booking modification with a subsequent cancellation; additional interaction patterns can make the request ambiguous without changing its underlying requirements.

Each blueprint is then instantiated as a concrete task with a provisional reference solution and tested through agent interaction. \emph{Verification} uses execution evidence and policy checks to assess the proposed reference and observed solution, guiding revision of either side when conflicts arise. Throughout refinement, the blueprint preserves what the task is intended to teach, connecting targeted behavioral coverage to verified examples.

\subsection{Compiling Environment Artifacts to Guarded Operators}

To make the environment's behavioral requirements reusable during synthesis, we ask an LLM to organize the supplied policies, tool implementations, and state schemas into a persistent library. Each entry centers on one atomic action and records its purpose, parameters, relevant state fields, guarded-condition branches and conditional consequences. We call this a \emph{guarded operator}:
\begin{equation}
g(a) = \{r_{a,i} \Rightarrow q_{a,i}\}_{i=1}^{m_a},
\end{equation}
where each guard $r_{a,i}$ combines applicable conditions using conjunction, disjunction, or negation, and $q_{a,i}$ describes the corresponding effects or required behavior, including state updates, follow-up actions, refusals, or failures. For example, a cancellation operator distinguishes eligible bookings that trigger refunds from ineligible requests requiring refusal.

These descriptions expose how an action's effects can establish or invalidate another action's conditions. They serve as an LLM-readable representation for composing behavioral requirements.

\subsection{Coverage-Guided Scenario Evolution}
\label{sec:composition}

\textsc{Gradus} uses the operator library to construct symbolic scenario blueprints. A blueprint specifies the user goal, relevant context, known and unknown information, and intended operator outcomes, while leaving concrete entities and database values unbound. Construction proceeds through the three stages below: at each stage, code governs the evolution schedule and samples the transformation instructions, and the LLM realizes each instruction into a contextually coherent blueprint.

\paragraph{Atomic blueprints.}
For each guarded operator, the LLM receives the full operator catalog and generates $k$ initial blueprints exercising its conditional behaviors. A shared coverage ledger tracks all the (policy, outcome) pairs of all operators represented in generated blueprints. Generation prompts include this ledger to prioritize underrepresented outcomes, both during atomic scenario generation and subsequent structural evolution.

\paragraph{Structural evolution.}
For each parent blueprint, code requests up to $K$ children by independently sampling a structural kernel for each attempt. It supplies the LLM with the parent, kernel instruction, operator catalog, and current coverage ledger. The kernel specifies how to extend the parent: \textsc{causal} connects actions through a state change, \textsc{parallel} adds an independent request, and \textsc{repetition} applies an action across a user-described set. The LLM chooses compatible operators and outcomes, retaining the parent's requirements while adding new ones, or declines an incoherent extension. For example, causal evolution may add a booking modification whose effects change the conditions of a subsequent cancellation. Repeating this procedure builds a lineage tree whose branching factor and depth control the breadth and extent of composition.

\paragraph{Interaction variation.}
After structural evolution, code samples interaction kernels and asks the LLM to rewrite each blueprint accordingly. These kernels introduce indirect references, ambiguity, user pressure, or additional reasoning while preserving the intended actions and outcomes. Thus, structural kernels change what the task requires, whereas interaction kernels change how those requirements will be surfaced and interpreted. New interaction dimensions can be added through additional kernel instructions.

Across all stages, the programmatic controller allocates search across structural kernels, and interaction variations, while the LLM performs model-guided pruning: it realizes an assigned expansion, selects a compatible alternative, or declines when the scenario-kernel combination is unrealistic. This avoids enumerating the full Cartesian product while preventing generation from collapsing onto the model's preferred patterns.

\subsection{Preserving Coverage through Task and Solution Validity}

A blueprint specifies what a task should teach, but its task, reference, and demonstration must still be constructed, with harder blueprints more likely to fail on the first attempt. \textsc{Gradus} therefore jointly refines both: \emph{task validity} concerns whether the grounded task is coherent and feasible, while \emph{solution validity} concerns whether a solution fulfills the task under the environment's policies. The two processes take each other's results as reciprocal feedback: Task design cross-checks simulated solution outcomes to diagnose task design defects and guide its next revision; solution search cross-checks the task and reference to produce reflection feedback for subsequent attempts. Environment execution anchors both directions.  The process follows an initialization $\rightarrow$ simulation $\rightarrow$ verification loop, while preserving the blueprint's behavioral requirements.

\paragraph{Initialization: ground blueprints to environment states.}
An LLM agent receives the blueprint, tools, policies, state schemas, and an environment-probing tool. It inspects the state and constructs a concrete \emph{task} and a candidate \emph{reference} action sequence. Code immediately replays the reference and returns execution errors for local revision. This first check establishes executability; whether the task is coherent and the reference fulfills it remains subject to cross-checking through simulation.

\paragraph{Simulation: task-guided solution refinement.}
\label{sec:simulation-search}
A user simulator receives the task and interacts with a solving agent that sees only the conversation and tool responses, never the hidden task specification or reference, and their interaction produces a full \emph{demonstration} candidate. Code launches $K$ parallel search branches with at most $m$ attempts each. A trajectory evaluator scores each attempt for action correctness, policy compliance, and communication quality, judging against the proposed task and reference, the policies, and the execution evidence, and returns feedback that guides the next attempts. The task and reference thus steer simulation reflection while remaining hidden from the solving agent. By the time an outcome reaches verification (next phase), parallel sampling and feedback-guided retries have already searched for the strongest demonstration the task admits.

\paragraph{Verification: demonstration-guided task refinement.}
This phase decides which (task, reference, demonstration) tuples to accept and, for the rest, arbitrates the refinement direction for the next loop: repair a flawed task, or keep searching for a better demonstration. To do so, code replays the reference and the demonstrated actions from the same initial state and compares their outcomes. Acceptance is decided by: when the two outcomes agree and the programmatic policy checks pass, the whole tuple is accepted as gold-standard supervision. When they do not, a diagnostic LLM reads the task, reference, demonstration, outcome differences, and check results and attributes the failure. Demonstration errors route the task back to further solution search, whereas evidence of a flawed task or reference (an incoherent or unsolvable task, hallucinated states, an incorrect reference or outcome) is summarized as feedback and returned to initialization. The language model only interprets the replay evidence and never certifies correctness itself, so a mutually agreeable but environment-inconsistent tuple cannot be accepted.

GRADUS thus treats task design and solution search as two provisional halves refined against each other, with execution in the environment as the shared arbiter that accepts a pair only when both hold. By recovering difficult blueprints rather than discarding them at their first failure, this loop carries the coverage produced by scenario evolution through to the training data, where the hardest lessons would otherwise be lost.

\section{Experiments}
\subsection{Experimental Setup}
\label{sec:setup}

\textbf{Environments and evaluation.}
We evaluate curriculum quality by downstream success on two customer-facing agent benchmarks: $\tau^2$-Bench airline, retail, and telecom domains~\citep{barres2025tau2benchevaluatingconversationalagents} and CAR-Bench~\citep{kirmayr-etal-2026-car}, an in-car voice assistant, with base, disambiguation, and hallucination task splits. Each task is scored by its benchmark's executable evaluator as a pass or fail, and per-split success is reported as pass@1 averaged over five independent runs, $\overline{\mathrm{pass}}@1=\tfrac{1}{5}\sum_{i=1}^{5}\mathbb{I}[y_i=1]$. Variance and consistency metrics are reported in Appendix~\ref{sec:main_result_details}. Benchmark tasks and references are held out from training curriculum construction, training, and checkpoint selection.

\textbf{Curriculum construction and training.}
GRADUS uses either Claude Opus 4.5~\citep{anthropic2025opus45} or Qwen3-235B-A22B-Thinking-2507~\citep{qwen2025qwen3235bthinking} for all model-mediated synthesis roles and trains Qwen3-8B~\citep{qwen2025qwen38b}, Qwen3-14B~\citep{qwen2025qwen314b}, and Qwen3-30B-A3B students~\citep{qwen2025qwen330bthinking}. Shallow lessons (evolution depths 1-3) are routed primarily to SFT, whereas deeper lessons (depths 4-5) are routed primarily to RL. Full-parameter SFT is followed by GRPO-based RL~\citep{shao2024deepseekmathpushinglimitsmathematical}, whose reward is the same executable acceptance signal used during construction, so RL optimizes verified targets rather than a learned or heuristic score.

\textbf{Model references.}
Table~\ref{tab:main} reports proprietary frontier models (Opus 4.5~\citep{anthropic2025opus45} and GPT-5.2~\citep{openai2025gpt52}), open-weight frontier-scale models: Kimi-K2-Thinking~\citep{moonshot2025kimik2thinking} and DeepSeek-V3.2~\citep{deepseek2025v32}, and off-the-shelf Qwen3 backbones, including Qwen3-235B-A22B~\citep{qwen2025qwen3235bthinking} as our open-weight proposal model. We also evaluate released 8B generalist-agent checkpoints to measure how agents broadly trained with generalist data synthesis recipes transfer to these policy-constrained environments~\citep{song-etal-2026-envscaler,wang2025klearagentforgeforgingagenticintelligence,sun2026agentskillerscalinggeneralistagent,lyu-etal-2026-agenticqwen}.

\textbf{Synthesis baselines.}
We train the same Qwen3-8B student on the released Simia~\citep{li2025simulatingenvironmentsreasoningmodels}, CoVe~\citep{chen2026covetraininginteractivetooluse}, and WRIT~\citep{gu2026writwritereadintensivetrajectory} datasets using our full-parameter SFT recipe, matching the number of tasks and maximum tokens per example across the baselines and our methods; these comparisons are limited to airline and retail, where their data are available. We further compare three paradigms under the same Opus 4.5 teacher, Qwen3-8B student, and data budget: \emph{Model-prior synthesis} provides the teacher with the environment policy, tools, and state schema but no explicit task structure or coverage guidance; the teacher freely proposes a blueprint, grounds it into a task and oracle, and retains it only when simulation produces the designed outcome. \emph{Graph-based synthesis} replaces free generation with a set of ontology constructed from tool graphs, each generating multiple blueprints before applying the same grounding and simulation-based rejection procedure. \emph{Trajectory-first synthesis} instead lets the LLM explore the live environment to obtain executable trajectories, uses a global tracker to discourage repeated action sequences, and derives a user scenario and oracle backward from each trajectory, followed by final replay. None of these baselines uses GRADUS's adaptive coverage expansion or verification-guided task recovery.

\subsection{Main Results}
\label{sec:results}

\begin{table}[t]
\caption{Task success on $\tau^2$-Bench and CAR-Bench, with the six-split average (Avg) at right. GRADUS (ours) rows are shaded; \textbf{bold} marks the best result among all rows outside the \emph{Frontier models} block. }
\label{tab:main}
\begin{center}
\small
\setlength{\tabcolsep}{5pt}
\begin{tabular}{ll ccc ccc c}
\toprule
& & \multicolumn{3}{c}{$\tau^2$-Bench} & \multicolumn{3}{c}{CAR-Bench} & \\
\cmidrule(lr){3-5}\cmidrule(lr){6-8}
Model & Proposer & Air & Ret & Tel & Base & Disamb & Halluc & Avg \\
\midrule
\multicolumn{9}{l}{\emph{Frontier models}} \\
Opus 4.5         & -- & $0.88$ & $0.89$ & $0.92$ & $0.85$ & $0.68$ & $0.64$ & $0.81$ \\
GPT-5.2          & -- & $0.82$ & $0.88$ & $0.89$ & $0.79$ & $0.52$ & $0.88$ & $0.80$ \\
Kimi-K2-Thinking & -- & $0.84$ & $0.88$ & $0.79$ & $0.56$ & $0.42$ & $0.67$ & $0.69$ \\
DeepSeek-V3.2    & -- & $0.78$ & $0.91$ & $0.76$ & $0.64$ & $0.48$ & $0.57$ & $0.69$ \\
\midrule
\multicolumn{9}{l}{\emph{Open-weight backbones}} \\
Qwen3-8B   & -- & $0.38$ & $0.41$ & $0.40$ & $0.40$ & $0.32$ & $0.40$ & $0.39$ \\
Qwen3-14B  & -- & $0.42$ & $0.60$ & $0.42$ & $0.49$ & $0.35$ & $0.42$ & $0.45$ \\
Qwen3-30B  & -- & $0.63$ & $0.77$ & $0.38$ & $0.52$ & $0.43$ & $0.45$ & $0.53$ \\
Qwen3-235B & -- & $0.61$ & $0.80$ & $0.50$ & $0.64$ & $0.46$ & $0.46$ & $0.58$ \\
\midrule
\multicolumn{9}{l}{\emph{Generalist agent training (released checkpoints)}} \\
EnvScaler (8B)        & -- & $0.36$ & $0.57$ & $0.33$ & $0.48$ & $0.34$ & $0.34$ & $0.40$ \\
Klear-AgentForge (8B) & -- & $0.39$ & $0.70$ & $0.26$ & $0.46$ & $0.29$ & $0.36$ & $0.41$ \\
AgentSkiller (8B)      & -- & $0.34$ & $0.64$ & $0.65$ & $0.49$ & $0.34$ & $0.33$ & $0.47$ \\
AgenticQwen (8B)      & -- & $0.44$ & $0.64$ & $0.45$ & $0.55$ & $0.32$ & $0.56$ & $0.49$ \\
\midrule
\multicolumn{9}{l}{\emph{Environment-targeted synthesis (matched student \& budget)}} \\
Simia (8B) & GPT-5/o4-mini & $0.36$ & $0.60$ & -- & -- & -- & -- & -- \\
CoVe (8B)  & Qwen3-235B    & $0.31$ & $0.66$ & -- & -- & -- & -- & -- \\
WRIT (8B)  & Qwen3.6-Plus  & $0.58$ & $0.78$ & -- & -- & -- & -- & -- \\
\midrule
\multicolumn{9}{l}{\emph{Environment-targeted synthesis (matched proposer, student \& budget)}} \\
Model-Prior Synthesis (8B)      & Opus-4.5 & $0.46$ & $0.53$ & $0.57$ & $0.40$ & $0.44$ & $0.31$ & $0.45$ \\
Graph-Based Synthesis (8B)      & Opus-4.5 & $0.53$ & $0.59$ & $0.62$ & $0.49$ & $0.36$ & $0.46$ & $0.51$ \\
Trajectory-First Synthesis (8B) & Opus-4.5 & $0.59$ & $0.70$ & $0.71$ & $0.56$ & $0.48$ & $0.52$ & $0.59$ \\
\midrule
\multicolumn{9}{l}{\emph{GRADUS (ours)}} \\
\rowcolor{blue!8} Qwen3-8B & Qwen3-235B & $0.71$ & $0.74$ & $0.64$ & $0.69$ & $0.45$ & $0.51$ & $0.62$ \\
  \addlinespace[1pt]
  \rowcolor{blue!8} Qwen3-8B & Opus-4.5 & $0.71$ & $0.85$ & $0.92$ & $0.76$ & $0.58$ & $0.74$ & $0.76$ \\
  \addlinespace[1pt]
  \rowcolor{blue!8} Qwen3-30B-A3B & Qwen3-235B & $0.69$ & $0.85$ & $0.75$ & $0.72$ & $0.60$ & $0.55$ & $0.69$ \\
  \addlinespace[1pt]
  \rowcolor{blue!8} Qwen3-30B-A3B & Opus-4.5 & $0.75$ & $0.86$ & $0.90$ & $0.80$ & $0.74$ & $0.74$ & $0.80$ \\
  \addlinespace[1pt]
  \rowcolor{blue!8} Qwen3-30B-A3B +RL & Opus-4.5 & $\mathbf{0.80}$ & $\mathbf{0.89}$ & $\mathbf{0.97}$ & $\mathbf{0.83}$ & $\mathbf{0.75}$ & $\mathbf{0.91}$ & $\mathbf{0.86}$ \\
\bottomrule
\end{tabular}
\end{center}
\end{table}

Table~\ref{tab:main} reports the main results across $\tau^2$-Bench and CAR-Bench.

\textbf{Curriculum construction matters under matched controls.}
With the Opus 4.5 proposer, Qwen3-8B student, and data budget fixed, GRADUS outperforms all three controlled synthesis baselines on every split. Compared with the strongest baseline, success increases from 59\% to 71\% on airline, from 71\% to 92\% on telecom, and from 52\% to 74\% on CAR hallucination. The resulting 8B student also surpasses Kimi-K2-Thinking and DeepSeek-V3.2 across all CAR-Bench splits. 
\textbf{GRADUS remains effective with a weaker proposer.}
Using Qwen3-235B and fewer than 1000 examples per domain, GRADUS raises the 8B student's average success from 40\% to 70\% on $\tau^2$-Bench and from 37\% to 55\% on CAR-Bench. It leads the generalist-agent baselines on four of six splits and outperforms Simia and CoVe on both available domains. The only environment-targeted exception is WRIT on retail (78\% versus 74\%), where WRIT uses a different synthesis model, Qwen3.6-Plus.
\textbf{The curriculum transfers across architectures and supports effective RL.} Without resynthesizing the curriculum, transferring from the dense Qwen3-8B student to the Qwen3-30B-A3B mixture-of-experts model maintains or improves performance on five of six splits, with the largest gain on CAR disambiguation (58\% to 74\%). Executable RL then improves all six splits, most notably increasing CAR hallucination success from 74\% to 91\%. The final model reaches 80\%, 89\%, and 97\% on the three $\tau^2$-Bench domains, matching Opus 4.5 on retail and achieving the best result on telecom. On CAR-Bench, its 83\% average surpasses that of every frontier model, while also achieving the best overall results on disambiguation and hallucination. Additional comparisons across 8B, 14B, and 30B student models are provided in Appendix Table~\ref{tab:scaling-tau} and Table~\ref{tab:scaling-car}.

\subsection{Ablations on Curriculum Components}
\label{sec:ablation}

Table~\ref{tab:ablation} isolates each curriculum component. Removing guarded operators lowers performance on both domains, as the proposer must then infer guards and outcomes from raw tool schemas, policies, and state definitions implicitly. The global ledger matters most for telecom, whose operators contain many policy-outcome pairs: corpus-wide coverage tracking is what steers the agent from skipping a required diagnostic step. Structural evolution matters as well, as it steers the proposer from overproducing familiar scenario structures. Interaction variation matters much more for CAR disambiguation, because varying how requests are surfaced demands different levels of interpretation, reasoning, and clarification. The validity ablations show a complementary effect. Removing verification causes a substantial drop, although the resulting curriculum remains useful because coverage-guided search can still produce difficult tasks whose solutions happen to succeed on the first attempt. Parallel search improves performance by combining outcome-based matching with multiple solution attempts. Adding policy-grounded reflection yields a particularly large gain on CAR disambiguation, where the proposer often fails on its first attempt but improves after receiving targeted feedback. Finally, task recovery uses simulation feedback to redesign failed tasks, preserving difficult yet valid examples and producing a more complete curriculum.

\begin{table}[h]
\caption{Ablation of GRADUS components with Qwen3-235B-A22B as proposal model and Qwen3-30B-A3B as student backbone. SFT task success (mean $\pm$ std) on $\tau^2$-Bench telecom and Car-Bench disambiguation. Left: leave-one-out ablation; right: cumulative validity loop.}

\label{tab:ablation}
\begin{center}
\small
\setlength{\tabcolsep}{4pt}
\begin{tabular}{@{}lcc@{\hspace{1.5em}}lcc@{}}
\toprule
Remove one component & Tel. & Dis. & Add validity cumulatively & Tel. & Dis. \\
\midrule
Full (Ours)                  & \ms{0.75}{0.02} & \ms{0.60}{0.05} & No validity                     & \ms{0.65}{0.05} & \ms{0.40}{0.04} \\
$-$ Guarded operators        & \ms{0.64}{0.03} & \ms{0.45}{0.03} & $+$ Solution: parallel   & \ms{0.68}{0.02} & \ms{0.43}{0.03} \\
$-$ Global ledger & \ms{0.54}{0.03} & \ms{0.52}{0.04} & $+$ Solution: reflection & \ms{0.71}{0.04} & \ms{0.55}{0.05} \\
$-$ Structural evolution  & \ms{0.63}{0.03} & \ms{0.55}{0.05} & $+$ Task recovery      (full)          & \ms{0.75}{0.02} & \ms{0.60}{0.05} \\
$-$ Interaction variation     & \ms{0.67}{0.02} & \ms{0.46}{0.04} &                    &  &  \\
\bottomrule
\end{tabular}
\end{center}
\end{table}

\subsection{Curriculum Coverage}
\label{sec:coverage}

\begin{figure}[h]
    \centering
    \includegraphics[width=\textwidth]{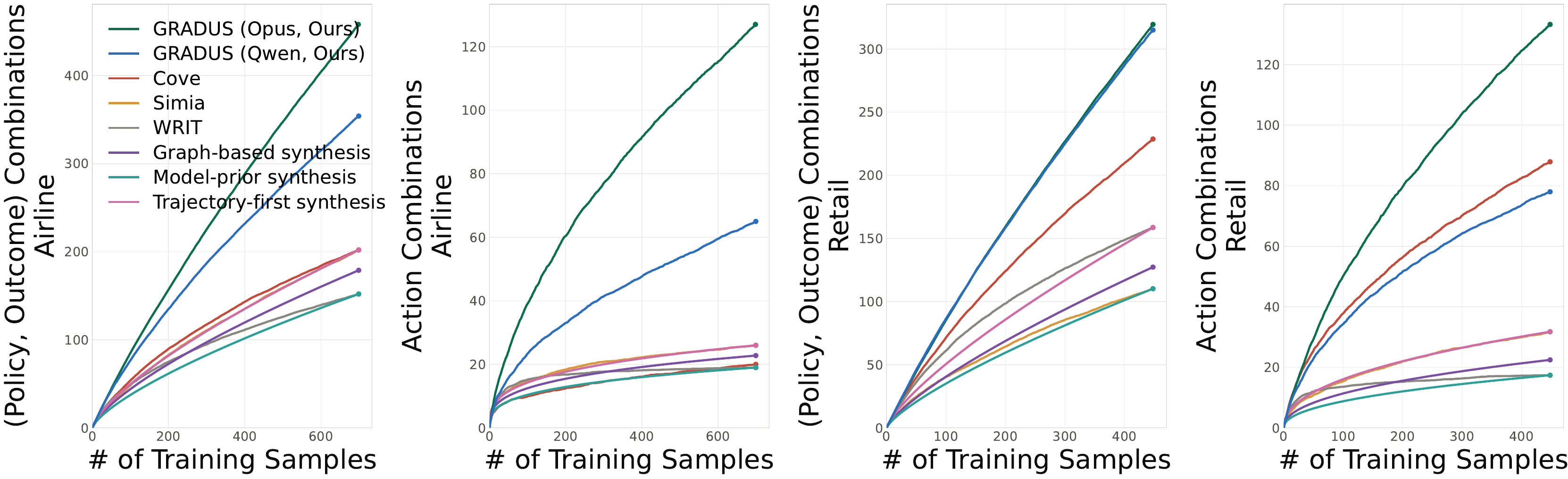}
    \caption{\textbf{Dataset Coverage by Environment Dimensions }(1,2: Airline; 3,4: Retail).
    \textbf{Panel-1,3}: Policy-Outcome Coverage. \textbf{Panel-2,4}: Action Combinations. }
    \label{fig:coverage}
\end{figure}

Because \textsc{Gradus} targets coverage gaps, it explores substantially more of the task space on both airline and retail (Figure~\ref{fig:coverage}). Panels~(1,3) measure distinct \emph{(policy, outcome) combinations}; coverage-guided search discovers $1.4\times$ to $2.9\times$ more combinations compared to baseline methods, exposing the student to a broader range of interacting policies and outcomes. Panels~(2,4) measure \emph{action combinations} among state-changing tools, where structural evolution expands coverage by up to ${\sim}5\times$ on airline and ${\sim}1.5\times$ on retail. In all four panels, baseline coverage plateaus as additional examples accumulate, indicating repeating familiar patterns, whereas \textsc{Gradus} continues discovering new ones. Proposer strength also matters: the two proposers coincide on retail (policy, outcome) coverage but differ by roughly $1.7\times$ on action combinations, indicating that composing unseen action sets depends on proposer capability more than enumerating policy branches does.

\subsection{From Intermittent Demonstrations to Reliable Learning}
\label{sec:stochastic}

\begin{wrapfigure}{r}{0.5\linewidth}
  \centering
  \includegraphics[width=1.0\linewidth]{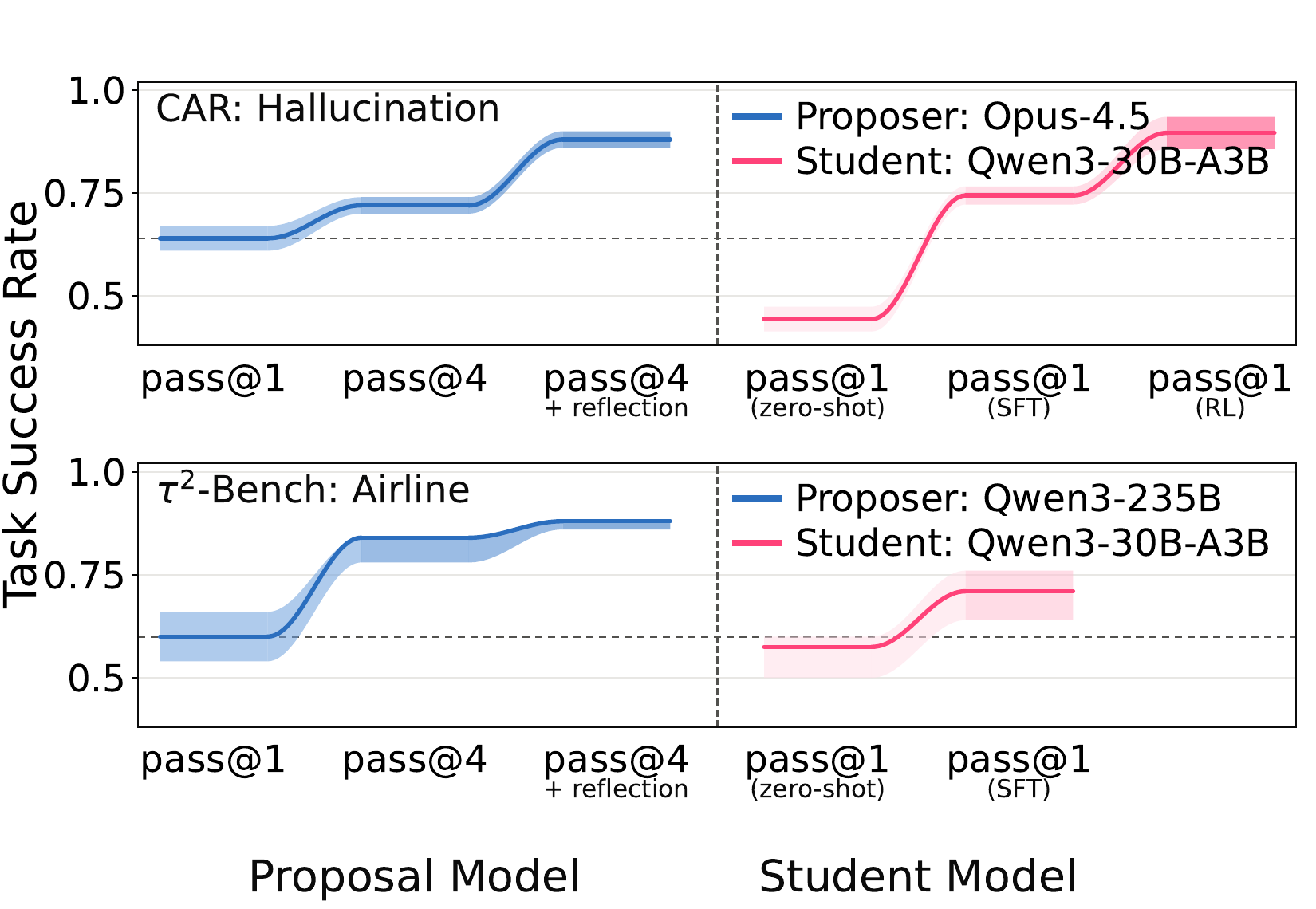}
  \caption{From intermittent frontier-model capability into reliable student performance. }
  \label{fig:reliability}
\end{wrapfigure}

As discussed in Section~\ref{sec:simulation-search}, solution generation is cast as a search problem; the two search strategies (parallel rollouts and reflection-guided retries) recover capability that a single reference-blind proposal model would miss at its first attempt (Figure~\ref{fig:reliability}). The two settings compare both proposal models: proprietary proposer (Opus 4.5 on CAR: Hallucination, top) and open-weight proposer (Qwen3-235B on airline, bottom). The proposer improves from 0.64 and 0.61 at pass@$1$ to 0.72 and 0.84 at pass@$4$, and further to 0.88 and 0.88 with evaluator-guided retries. Training then converts this intermittent capability into reliable single-shot behavior: SFT lifts the student from 0.45 to 0.74 and from 0.63 to 0.69, surpassing both proposers' pass@$1$. RL raises this further to 0.91 on CAR-Bench, exceeding the proposer's observed search success rate of 0.88.

\subsection{How Curriculum Preservation Retains Difficult Lessons}
\label{sec:preservation}

\begin{figure*}[h]
  \centering
  \includegraphics[width=\textwidth]{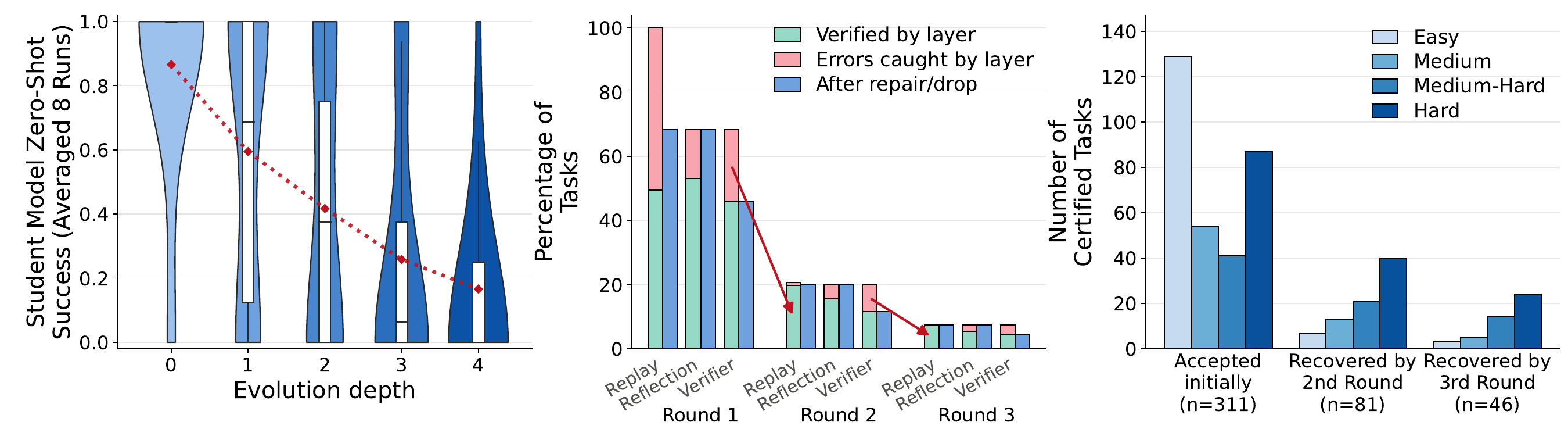}
  \caption{Curriculum preservation retains difficult lessons. \textbf{(a)} Student zero-shot success (averaged over 8 runs) versus evolution depth. \textbf{(b)} Replay, reflection and recovery dynamics per loop. \textbf{(c)} Verified tasks grouped by the round in which they were accepted, split by difficulty class (Easy/Medium/Medium-Hard/Hard).}
  \label{fig:preservation}
\end{figure*}

Figure~\ref{fig:preservation} shows how iterative recovery preserves curriculum difficulty. In the left panel, mean zero-shot success of the off-the-shelf student model falls monotonically from $0.87$ to $0.17$ with evolution depth, demonstrating controlled difficulty growth. The middle panel traces the full initialization $\rightarrow$ simulation $\rightarrow$ verification loop, with the red arrows marking tasks to be re-initialized in the next round. This process is not a filtering funnel: the final curriculum accumulates the post-verification outputs from every round (all blue bars), rather than retaining only Round-3. The right panel shows that recovery disproportionately retains difficult examples: medium-hard and hard tasks comprise 41\% of initially accepted tasks but 75\% and 83\% of those recovered in Rounds-2 and-3. By recovering rather than discarding difficult proposals, the loop preserves the difficulty introduced by coverage-guided search and supports the downstream results in Table~\ref{tab:main}.

\subsection{From Curriculum to Frontier Exams}
\label{sec:stress}

Beyond synthesizing training data, we investigate whether \textsc{Gradus} could construct frontier exams for the proposer, since the environment is cast as one latent syllabus for both: coverage-guided search targets uncovered \emph{(policy, outcome)} and action combinations, which initialization, simulation, and verification realize harder tasks. Along similar lines, TASTE~\citep{keren2026mattertasteimprovingcoverage} searches tool-name sequences and falls back to simpler variants when harder evolution fails; \textsc{Gradus} recovers the evolved task and its solution while preserving the difficulty requirements. With a frontier model as scout, this process therefore locates supported regions that the frontier model solves only intermittently.
Using Opus~4.5 as scout, we evolve airline, retail, and CAR-Bench base tasks to depth five and retain verified tasks whose gold outcome is matched in at most five of eight rollouts. Two annotators audited all 58 $\tau^2$-Bench and 79 CAR-Bench candidates, judging $87.5\%$, $100\%$, and $90.5\%$ correct, with Gwet's agreement coefficient AC1~\citep{gwet2008computing} of $0.79$, $1.00$, and $0.96$ (Appendix~\ref{app:annotation}).
On tasks both annotators accepted as correct, Opus-4.5 (proposal model) falls from $88\%$, $89\%$, and $85\%$ to $52\%$, $60\%$, and $43\%$, and three other  held-out models decline by $18\%$ to $44\%$ in every domain (Table~\ref{tab:evolved}), so the difficulty transfers to models held out from the exam construction process.

\begin{table}[h]
\caption{Task success on the original benchmarks and the frozen frontier exams (mean $\pm$ std over five runs). Results are reported on $\tau^2$-Bench (Airline and retail), and CAR-bench (base). Opus 4.5 serves as the scout; all other models are held out during the evolved exam synthesis.}
\label{tab:evolved}
\begin{center}
\small
\setlength{\tabcolsep}{4pt}
\begin{tabular}{lcccccc}
\toprule
 & \multicolumn{2}{c}{Airline ($\tau^2$)} & \multicolumn{2}{c}{Retail ($\tau^2$)} & \multicolumn{2}{c}{Base (CAR)} \\
\cmidrule(lr){2-3} \cmidrule(lr){4-5} \cmidrule(lr){6-7}
Model & Original & Evolved & Original & Evolved & Original & Evolved \\
\midrule
GPT-5.2 & \ms{0.82}{0.05} & \ms{0.64}{0.08} & \ms{0.88}{0.02} & \ms{0.62}{0.08} & \ms{0.79}{0.02} & \ms{0.44}{0.04} \\
Opus 4.5 & \ms{0.88}{0.01} & \ms{0.52}{0.08} & \ms{0.89}{0.01} & \ms{0.60}{0.02} & \ms{0.85}{0.02} & \ms{0.43}{0.03} \\

DeepSeek-V3.2 & \ms{0.78}{0.03} & \ms{0.43}{0.08} & \ms{0.91}{0.02} & \ms{0.57}{0.04} & \ms{0.64}{0.02} & \ms{0.26}{0.04} \\
Kimi-K2-Thinking & \ms{0.84}{0.03} & \ms{0.40}{0.10} & \ms{0.88}{0.01} & \ms{0.49}{0.16} & \ms{0.56}{0.03} & \ms{0.17}{0.03} \\

\bottomrule
\end{tabular}
\end{center}
\end{table}

\section{Conclusion}
\label{sec:conclusion}

We introduced \textsc{Gradus}, a framework that treats a fixed executable environment as a latent syllabus and separates the behaviors to be taught from the tasks and solutions that teach them. \textsc{Gradus} organizes environment logic into guarded operators, uses a global coverage ledger to steer composition toward underrepresented policy outcomes and action combinations, and refines tasks and solutions through reciprocal feedback, so verification recovers difficult lessons rather than filtering them out. Under matched controls, the resulting curricula outperform alternative synthesis methods, transfer across student architectures, and support executable RL: the final 30B-A3B model reaches $89\%$ average success on $\tau^2$-Bench and $83\%$ on CAR-Bench, surpassing all evaluated frontier model on the latter. The same machinery turns advanced syllabus regions into frontier exams that remain challenging for held-out frontier models. Thus, the agent-data synthesis could be framed as controlled curriculum search, requiring control over both which behaviors are explored and which survive verification. Extending this machinery beyond bounded environments with explicit policies and executable outcomes remains important future work.

\bibliography{iclr2027_conference}
\bibliographystyle{iclr2027_conference}

\appendix
\section{Appendix}

\subsection{Related Work}
\label{sec:related}

\textbf{Scaling data and environments for generalist agents.}
A growing line of work scales agent training by synthesizing executable environments together with their tasks, rewards, or interaction trajectories~\citep{fang-etal-2026-towards,cai2025autoforgeautomatedenvironmentsynthesis,tu2026scaleenvscalingenvironmentsynthesis,song-etal-2026-envscaler,wang2026cuagymscalingverifiabletraining}. Complementary efforts construct broad tool-use corpora spanning many interfaces and domains, including ToolMind~\citep{yang2025toolmindtechnicalreportlargescale}, GEM~\citep{xu-etal-2026-unlocking}, AgentSkiller~\citep{sun2026agentskillerscalinggeneralistagent}, AgenticQwen~\citep{lyu-etal-2026-agenticqwen}, Klear-AgentForge~\citep{wang2025klearagentforgeforgingagenticintelligence}, and TOUCAN~\citep{xu2025toucansynthesizing15mtoolagentic}. These methods primarily seek breadth and transfer across many worlds. \textsc{Gradus} addresses the complementary problem of achieving systematic behavioral depth within a fixed target environment: it seeks systematic depth within that environment by expanding coverage over its guarded operations and their compositions, and by validating every lesson through deterministic replay in the system itself.

\textbf{Diverse and verifiable task synthesis for interactive agents.}
Existing methods obtain diversity through seed amplification or blueprint generation~\citep{li2025simulatingenvironmentsreasoningmodels,prabhakar2025apigenmtagenticpipelinemultiturn}, explicit constraints or predefined dimensions~\citep{chen2026covetraininginteractivetooluse,gu2026writwritereadintensivetrajectory}, domain-entity-graph traversal~\citep{sun2026agentskillerscalinggeneralistagent}, iterative workflow expansion~\citep{lyu-etal-2026-agenticqwen}, or execution-first exploration~\citep{wang2026trajectory2tasktrainingrobusttoolcalling,chen2026divescalingdiversityagentic,zeng2026logigenlogicdrivengenerationverifiable}, and assess solvability by exact state equivalence~\citep{zeng2026logigenlogicdrivengenerationverifiable}, state-difference and action matching~\citep{sun2026agentskillerscalinggeneralistagent}, generated programmatic checkers~\citep{song-etal-2026-envscaler}, constraint checklists~\citep{chen2026covetraininginteractivetooluse}, or LLM rubrics~\citep{lyu-etal-2026-agenticqwen}; others transform tasks under instance-specific or oracle-preserving checks~\citep{gao2026selfevolvingsyntheticdataverifiablereward,xu2026controllableverifiabletoolusedata}. \textsc{Gradus} differs in what it synthesizes and how it is controlled. It makes the system's \emph{guarded operations} (policy-conditioned actions and their contrasting outcomes) the reusable unit of synthesis, and expands them within operations, across operations and into cross-space interaction challenges with ascending difficulty, a space broader than fixed dimensions, tool-graph paths, or behavior-tree branches. An adaptive coverage ledger steers generation toward under-covered targets globally, while the task verification is decided by executable environment replay with the model diagnosing rather than judging, and a selected lesson that first fails is recovered through task revision or expanded demonstration search instead of being discarded.

\label{app:comparison}
Table~\ref{tab:comparison} compares related methods along the design choices central to GRADUS: the source of candidate lessons, adaptive corpus-level
  coverage, the treatment of candidates that fail validation, added interaction challenges, and construction in or outside the target system. The comparison
  concerns curriculum design rather than overall system capability; partial marks denote mechanisms that apply only to a subset of data or operate at a different
  level, as detailed below.

 \begin{table}[h]
  \caption{Curriculum-design comparison. \cmark{} present; \pmark{} partial; \xmark{} absent. \textbf{Direct} and \textbf{Xfer} indicate construction in the target system and transfer from other systems.}
  \label{tab:comparison}
  \centering
  \footnotesize
  \setlength{\tabcolsep}{2.5pt}
  \renewcommand{\arraystretch}{1.08}
  \resizebox{\linewidth}{!}{%
  \begin{tabular}{@{}llc ccc ccc c@{}}
  \toprule
  & & & \multicolumn{3}{c}{Failed-lesson handling}
  & \multicolumn{3}{c}{Interaction challenges} & \\
  \cmidrule(lr){4-6}\cmidrule(lr){7-9}
  Method & Source & \shortstack{Adaptive\\coverage}
  & Drop & Revise & Demo
  & Human & Reason. & Cross-dom. & Transfer \\
  \midrule
  Simia~\citep{li2025simulatingenvironmentsreasoningmodels}
  & Seed & \xmark & \pmark & \xmark & \xmark
  & \xmark & \xmark & \xmark & Xfer \\

  CoVe~\citep{chen2026covetraininginteractivetooluse}
  & System & \xmark & \cmark & \xmark & \xmark
  & \pmark & \xmark & \xmark & Direct \\

  WRIT~\citep{gu2026writwritereadintensivetrajectory}
  & System & \xmark & \cmark & \xmark & \xmark
  & \cmark & \pmark & \xmark & Direct \\

  AgentSkiller~\citep{sun2026agentskillerscalinggeneralistagent}
  & World & \xmark & \cmark & \pmark & \xmark
  & \pmark & \xmark & \cmark & Xfer \\

  EnvScaler~\citep{song-etal-2026-envscaler}
  & World & \xmark & \pmark & \xmark & \xmark
  & \xmark & \xmark & \xmark & Xfer \\

  Klear-AgentForge~\citep{wang2025klearagentforgeforgingagenticintelligence}
  & Mixed & \xmark & \pmark & \xmark & \pmark
  & \xmark & \xmark & \xmark & Xfer \\

  AgenticQwen~\citep{lyu-etal-2026-agenticqwen}
  & Seed & \xmark & \cmark & \xmark & \xmark
  & \pmark & \xmark & \xmark & Xfer \\

  LOGIGEN~\citep{zeng2026logigenlogicdrivengenerationverifiable}
  & World & \xmark & \cmark & \pmark & \xmark
  & \pmark & \xmark & \xmark & Xfer \\
  \midrule
  GRADUS (ours)
  & System & \cmark & \pmark & \cmark & \cmark
  & \cmark & \cmark & \cmark & Direct \\
  \bottomrule
  \end{tabular}%
  }
  \end{table}

\subsection{Problem Formulation}
\label{app:formulation}

\paragraph{Agent-environment interaction.}
We study an agent \(\pi_\theta\) that communicates with a user and acts in a stateful tool environment. Given a user policy \(\mu\), the interaction with the environment from the agent's perspective is a partially observable Markov decision process
\begin{equation}
\mathcal M_\mu(E)
=
\bigl(
\mathcal X,
\mathcal A_{\mathrm{ag}},
\mathcal O_{\mathrm{ag}},
P_\mu,
Z_\mu,
\mathsf r,
\rho_0,
\gamma
\bigr).
\label{eq:agent-pomdp}
\end{equation}
The hidden state \(x_t\in\mathcal X\) includes the backend state, task state, and information private to the user. An agent action \(a_t\in\mathcal A_{\mathrm{ag}}\) is either a message or a tool call, while its observations include user messages and tool responses. The agent chooses actions from the interaction history \(h_t\) according to \(\pi_\theta(a_t\mid h_t)\). Its standard objective is
\begin{equation}
J_E(\pi_\theta;\mu)
=
\mathbb E_{\tau\sim(\pi_\theta,\mu,E)}
\left[
\sum_{t=0}^{H}\gamma^t \mathsf r_t
\right],
\label{eq:agent-objective}
\end{equation}
where the reward measures successful task completion under the environment's operating policies. 

\paragraph{Fixed executable environment.}
The agent-environment interaction process should be distinguished from the fixed environment specification consumed by \textsc{Gradus}. We represent this specification as
\begin{equation}
E
=
\bigl(
\mathcal S,
\mathcal A_{\mathrm{tool}},
\mathcal P_{\mathrm{op}},
F
\bigr),
\label{eq:system-specification}
\end{equation}
where \(\mathcal S\) is the backend state space described by the state schema, \(\mathcal A_{\mathrm{tool}}\) is the set of parameterized tool actions, \(\mathcal P_{\mathrm{op}}\) is the written operating policy, and
\[
F:\mathcal S\times\mathcal A_{\mathrm{tool}}
\rightarrow
\mathcal S\times\mathcal O_{\mathrm{tool}}
\]
is the executable tool semantics that returns a new state and tool response. For example, in the airline domain, \(\mathcal S\) contains reservation and payment records, \(\mathcal A_{\mathrm{tool}}\) includes actions such as \texttt{cancel\_reservation}, and \(\mathcal P_{\mathrm{op}}\) specifies when that action should be executed, refused, or preceded by additional steps.

The environment specification \(E\) is fixed: \textsc{Gradus} neither generates nor modifies the definitions of states, tools, policies, or implementation. Combined with a task and user policy \(\mu\), however, \(E\) induces the environment interaction process \(\mathcal M_\mu(E)\) in which the agent acts.

\paragraph{Budgeted Curriculum Construction}
\label{app:coverage-objective}
Equation~\ref{eq:agent-objective} assumes a task distribution; the role of \textsc{Gradus} is to construct one with broad behavioral coverage. Let \(\mathcal U_E\) contain the coverage units derived from \(E\), including atomic blueprints, structural evolutions, and interaction variations. Each blueprint \(z\) covers a subset \(\phi(z)\subseteq\mathcal U_E\). For a curriculum \(\mathcal D\), define
\begin{equation}
n_{\mathcal D}(u)
=
\sum_{d\in\mathcal D}
\mathbb I[u\in\phi(z_d)]
\end{equation}
as the number of examples covering unit \(u\).

We formulate curriculum construction as
\begin{equation}
\begin{aligned}
\max_{\mathcal D}\quad
& F(\mathcal D)
 =\sum_{u\in\mathcal U_E}
   \upsilon\!\left(n_{\mathcal D}(u)\right)\\
\text{subject to}\quad
& C(\mathcal D)\le B,\\
& V_E(d)=1,\qquad \forall d\in\mathcal D,
\end{aligned}
\label{eq:coverage-objective}
\end{equation}
where \(\upsilon\) is increasing with diminishing returns, \(C(\mathcal D)\le B\) imposes the data or construction budget, and \(V_E\) requires every retained task and solution to pass environment-grounded validation.

\textsc{Gradus} approximates this objective rather than solving it exactly. Its coverage ledger prioritizes underrepresented units, while joint task-solution refinement helps selected behaviors survive construction and enter the verified curriculum. Coverage serves as a controllable data-design objective; its effect on the trained policy is evaluated empirically.

\paragraph{Supervised fine-tuning.}
SFT establishes the behavioral prior: it teaches the student the tool-calling format, the
turn-taking discipline of the environment, and the policy-compliant action sequences that the
validity loop has already certified, giving RL a starting policy whose success rate is high
enough for group-relative comparisons to be informative. Given a verified curriculum
$\mathcal{D}_{\mathrm{SFT}}=\{(x_i,y_i)\}_{i=1}^{N}$, where $x_i$ contains the task and
interaction context and $y_i$ is the verified assistant trajectory, we optimize the standard
autoregressive negative log-likelihood:
\begin{equation}
\mathcal{L}_{\mathrm{SFT}}(\theta)
=
-\frac{1}{N}\sum_{i=1}^{N}
\frac{1}{\sum_t m_{i,t}}
\sum_{t=1}^{|y_i|}
m_{i,t}\log
\pi_\theta\!\left(y_{i,t}\mid x_i,y_{i,<t}\right),
\end{equation}
where $m_{i,t}\in\{0,1\}$ masks non-assistant tokens. Thus, loss is computed only over the
target assistant responses while the complete preceding interaction remains available as
context.

\paragraph{Reinforcement learning with GRPO.}
Starting from the SFT checkpoint,
we apply Group Relative Policy Optimization (GRPO). For each task $x$, the old policy
$\pi_{\mathrm{old}}$ samples a group of $G$ trajectories $\{y_i\}_{i=1}^{G}$ by interacting with
the live environment, which receive executable rewards $\{r_i\}_{i=1}^{G}$ from the same
environment-grounded checks used during verification. GRPO forms group-normalized advantages
$\hat{A}_i=(r_i-\bar r)/(\operatorname{std}(r)+\delta)$, with $\delta$ a small constant for
numerical stability, and optimizes
\begin{equation}
\mathcal{J}_{\mathrm{GRPO}}(\theta)
=
\mathbb{E}_{x,\{y_i\}}
\left[
\frac{1}{G}\sum_{i=1}^{G}\frac{1}{|y_i|}
\sum_{t=1}^{|y_i|}
\min\!\left(
\rho_{i,t}\hat A_i,\,
\operatorname{clip}(\rho_{i,t},1-\epsilon_{\mathrm{low}},
1+\epsilon_{\mathrm{high}})\hat A_i
\right)
\right],
\end{equation}
where $\rho_{i,t}=\pi_\theta(y_{i,t}\mid x,y_{i,<t})/\pi_{\mathrm{old}}(y_{i,t}\mid x,y_{i,<t})$
and the token index $t$ again ranges only over agent-generated tokens.

\subsection{Training Details}
\label{app:training}

\paragraph{SFT configuration.}
We fine-tune all parameters for three epochs with learning rate $1.0\times10^{-5}$ under a
cosine schedule with warmup ratio $0.1$ and maximum sequence length $32{,}768$, instantiating
the mask $m_{i,t}$ so that loss falls on the final assistant turn of each multi-turn example.
The checkpoint with the lowest loss on a held-out $5\%$ validation split is selected and, for
the RL stage, initializes the policy. This recipe is held fixed throughout: the same
optimization settings, sequence length, epoch count, and checkpoint-selection rule are applied
to every student size in Table~\ref{tab:scaling-tau} and Table~\ref{tab:scaling-car} and to every baseline curriculum in
Table~\ref{tab:main}, with no per-method or per-size tuning. Differences in downstream success
therefore reflect the training data under comparison rather than differences in how each
dataset was consumed.

\paragraph{RL configuration.}
RL uses GRPO advantage estimation with DAPO-style refinements~\citep{yu2025dapoopensourcellmreinforcement}: asymmetric clip thresholds
$\epsilon_{\mathrm{low}}=0.2$ and $\epsilon_{\mathrm{high}}=0.25$, token-level loss
aggregation, and KL regularization with coefficient $0.01$ under a low-variance estimator.
We train with learning rate $1\times10^{-6}$, batch size $8$, and $G=16$ rollouts per task,
under a budget of 32k length per rollout. Rollouts are collected by
interacting with the live environment, so every reward reflects an executed outcome.

\subsection{Algorithm}
  Algorithm~\ref{alg:gradus} follows the three stages in
  Equation~\ref{eq:GRADUS-overview}: constructing guarded operators, searching
  for lesson blueprints, and realizing them as verified training examples. The
  algorithm marks whether each operation is performed by an LLM or by code. Code
  controls search and execution, while the LLM constructs the semantic content.

\begin{algorithm}[h]
\caption{\textsc{Gradus}: coverage-guided curriculum construction from a fixed executable system.}
\label{alg:gradus}
\begin{algorithmic}[1]
\Require system $E=(\mathcal S,\mathcal A,\mathcal P,T)$; blueprints per operator $k$; children $K$; depth $D$; rollouts $R$; attempts $m$; budget $B$
\Ensure curriculum $\mathcal D$ of accepted $(\text{task},\text{reference},\text{demonstration})$ tuples
\State \textbf{Stage 1: Construct guarded operators}
\State $\mathcal G \gets \{\,\textsc{ConstructOperator}(a,\mathcal P,\textsc{Code}(a),\textsc{Schema}(a)) : a \in \mathcal A\,\}$ \Comment{\textsc{llm}: branches $r_{a,i}\!\Rightarrow\! q_{a,i}$, touched state, failures}
\State \textbf{Stage 2: Coverage-guided lesson search} \quad ($\mathcal Z \gets \varnothing$; ledger $L \gets \mathbf 0$)
\For{each $g(a) \in \mathcal G$}
  \State $\mathcal Z \gets \mathcal Z \cup \{\,k \text{ atomic blueprints } \textsc{Atomic}(g(a),\mathcal G,L)\,\}$; \; $L \gets \textsc{UpdateLedger}(L,\mathcal Z)$ \Comment{\textsc{llm} generates; \textsc{code} records $(a,\omega)$}
\EndFor
\For{depth $d = 1 \dots D$} \Comment{structural expansion}
  \For{each parent $z$ at depth $d-1$}
    \For{$c = 1 \dots K$}
      \State $z' \gets \textsc{Expand}(z,\kappa,\mathcal G,L)$, \; $\kappa \sim \{\text{causal},\text{parallel},\text{repetition}\}$ \Comment{\textsc{code} samples $\kappa$; \textsc{llm} adds operator or \textsc{decline}s}
      \If{$z' \neq \textsc{decline}$} $\mathcal Z \gets \mathcal Z \cup \{z'\}$ \EndIf
    \EndFor
    \State $L \gets \textsc{UpdateLedger}(L,\mathcal Z)$ \Comment{\textsc{code}: update coverage after this parent's children}
  \EndFor
\EndFor
\For{each $z \in \mathcal Z$}
  \State $z \gets \textsc{Decorate}(z,\sigma)$, \; $\sigma \sim \mathcal K_{\text{sem}}$ \Comment{\textsc{code} samples $\sigma$; \textsc{llm} rewrites to same actions, added surface}
\EndFor
\State \textbf{Stage 3: Realization with the validity loop} \quad ($\mathcal D \gets \varnothing$)
\For{each blueprint $z \in \mathcal Z$}
  \State $b \gets 0$; \; $\textit{fb} \gets \varnothing$
  \Repeat
    \State $(\textit{task},\textit{ref},s_0) \gets \textsc{Initialize}(z,E,\textsc{Probe},\textit{fb})$ \Comment{\textsc{llm}: probe state, ground task and reference}
    \If{$\neg\,\textsc{Replay}(\textit{ref},s_0)$} $\textit{fb} \gets \textit{err}$; \; $b \gets b+1$; \; \textbf{continue} \EndIf \Comment{\textsc{code}: reference pre-check}
    \State $\hat d \gets \textsc{Simulate}(\textit{task},\pi_{\text{blind}},U)$ \Comment{\textsc{llm}: $R$ rollouts $\times$ $m$ evaluator-guided retries}
    \State $\textit{match} \gets \textsc{ReplayCompare}(\textit{ref},\hat d,s_0)$; \; $\textit{pol} \gets \textsc{PolicyCheck}(\hat d,\mathcal P)$ \Comment{\textsc{code}: cross-examine by execution}
    \If{$\textit{match} \wedge \textit{pol}$}
      \State $\mathcal D \gets \mathcal D \cup \{(\textit{task},\textit{ref},\hat d)\}$; \; \textbf{break} \Comment{accept gold-standard tuple}
    \Else
      \State $\textit{side} \gets \textsc{Diagnose}(\textit{task},\textit{ref},\hat d,\textit{diff},\textit{checks})$ \Comment{\textsc{llm}: attribute defect}
      \State $\textit{fb} \gets$ feedback to \textsc{Initialize} if $\textit{side}=\textsc{task}$, else to \textsc{Simulate}
    \EndIf
    \State $b \gets b+1$
  \Until{accepted \textbf{or} $b \ge B$}
\EndFor
\State \Return $\mathcal D$ \Comment{route by depth: shallow $\to$ SFT, deep $\to$ RL}
\end{algorithmic}
\end{algorithm}

\subsection{Training Data}
\label{app:data}
\paragraph{Data statistics.}
Table~\ref{tab:datastats} reports the size of the final curriculum per domain
and proposer, and Table~\ref{tab:datastats-depth} breaks each total down by
evolution depth. SFT rows are per-assistant-step rows expanded from the
SFT tasks; RL items are one per task. Shallow lessons (depths 1-3) supply most
SFT tasks and deep lessons (depths 4-5) most RL tasks. 

\begin{table}[h]
\centering
\small
\setlength{\tabcolsep}{6pt}
\begin{tabular}{@{}lcccccc@{}}
\toprule
 & \multicolumn{3}{c}{Opus 4.5 proposer} & \multicolumn{3}{c}{Qwen3-235B proposer} \\
\cmidrule(lr){2-4}\cmidrule(lr){5-7}
Domain & SFT tasks & SFT rows & RL tasks & SFT tasks & SFT rows & RL tasks \\
\midrule
Airline      & 698   & 8{,}468  & 341   & 699 & 6{,}848 & 267 \\
Retail       & 585   & 7{,}971  & 381   & 448 & 5{,}096 & 249 \\
Telecom      & 414   & 6{,}145  & 367   & 370 & 5{,}162 & 168 \\
CAR (combined) & 1{,}901 & 11{,}091 & 1{,}227 & 986 & 5{,}117 & 1{,}018 \\
\bottomrule
\end{tabular}
\caption{Final curriculum size per domain and proposer. SFT tasks and RL tasks
are verified task--reference pairs; a task supplies either supervised
demonstrations or an RL reference. }
\label{tab:datastats}
\end{table}

\begin{table}[h]
\centering
\small
\setlength{\tabcolsep}{5pt}
\begin{tabular}{@{}lcccc@{}}
\toprule
Depth & Airline & Retail & Telecom & CAR (combined) \\
\midrule
\multicolumn{5}{@{}l}{\emph{Opus 4.5 proposer}} \\
1     & 156   & 128 & 178 & 718 \\
2     & 352   & 278 & 363 & 900 \\
3     & 267   & 223 & 192 & 660 \\
4     & 156   & 190 & 48  & 458 \\
5     & 108   & 147 & --  & 392 \\
Total & 1{,}039 & 966 & 781 & 3{,}128 \\
\midrule
\multicolumn{5}{@{}l}{\emph{Qwen3-235B proposer}} \\
1     & 466 & 162 & 139 & 605 \\
2     & 314 & 266 & 208 & 480 \\
3     & 147 & 187 & 141 & 410 \\
4     & 39  & 82  & 50  & 509 \\
Total & 966 & 697 & 538 & 2{,}004 \\
\bottomrule
\end{tabular}
\caption{Verified curriculum tasks by evolution depth. Each entry is the total
number of accepted tasks at that depth (supervised and RL tasks combined). The Opus proposer
evolves to depth $4$ on airline, retail, and CAR; telecom saturates at depth
$4$. The Qwen proposer is reported over
depths $1$-$4$. CAR combines the base,
disambiguation, and hallucination splits.}
\label{tab:datastats-depth}
\end{table}

\subsection{Ablation Details}
  \label{app:ablation_details}

  All ablations fix the proposer, student, data budget, and training procedure. \textbf{$-$ Guarded operators} replaces them with raw policies, tool schemas, and
  state schemas, leaving the proposer to infer constraints and scenarios. \textbf{$-$ Global ledger} retains evolution but removes corpus-level \emph{(policy,
  outcome)} tracking and gap selection. \textbf{$-$ Structural evolution} instead generates scenarios in one shot, without structural kernels or iterative
  expansion. \textbf{$-$ Interaction variation} sends structurally evolved scenarios directly to task realization. \textbf{No validity} trains on the first
  rollout without verification. \textbf{$+$ Solution: parallel} samples \(K\) rollouts and accepts one whose outcome matches the reference. \textbf{$+$ Solution:
  reflection} additionally gives each rollout up to two evaluator-guided retries based on policy compliance, outcome agreement, and communication quality.
  \textbf{$+$ Task recovery (full)} : task- or reference-side errors trigger re-initialization, while solution-side errors trigger
  further demonstration search, until verification succeeds or the recovery budget is exhausted.

\subsection{End-to-End Example: Evolving and Recovering a Policy-Interaction Lesson}
\label{app:end_to_end_example}

This section traces one airline example through guarded-operator extraction, structural evolution, task initialization, reflection-guided simulation, and task-side recovery. The artifacts below are lightly abridged for readability; all policy conditions, task requirements, and state-changing actions are preserved.

\paragraph{Latent lesson from guarded operators.}
The extracted operators reveal a causal interaction that is not apparent from the tool names alone. An old economy reservation without insurance cannot be cancelled for a change of plans. However, changing its cabin to business makes the business-cabin cancellation guard true.
\begin{artifactbox}{Guarded-operator artifact (abridged raw specification)}
  \small
  \emph{Source citations and low-level state paths are omitted.}

  \texttt{\textbf{action\_spec: cancel\_reservation}}
  \begin{itemize}
      \setlength{\itemsep}{2pt}
      \item \texttt{description:} Cancel the whole reservation.
      \item \texttt{parameters:}
            \texttt{reservation\_id: string, required}.
      \item \texttt{returns:} the updated reservation.

      \item \texttt{precondition: AND}
      \begin{itemize}
          \item obtain the user ID, reservation ID, and cancellation reason;
          \item list the cancellation and refund details and obtain explicit
                confirmation;
          \item verify that the reservation exists;
          \item verify that no flight has status
                \texttt{flying} or \texttt{landed};
          \item verify that at least one cancellation trigger holds:
          \[
          \begin{aligned}
          &\texttt{created\_within\_24h}\\
          {}\lor{}&\texttt{airline\_cancelled\_flight}\\
          {}\lor{}&\texttt{cabin=business}\\
          {}\lor{}&
          (\texttt{insurance=yes}\land
          \texttt{covered\_reason}).
          \end{aligned}
          \]
          \item if a certificate was used, disclose that its remaining value is
                not refundable.
      \end{itemize}

      \item \texttt{effects:}
      \begin{itemize}
          \item append a negative refund entry for every original payment;
          \item set \(\texttt{reservation.status}\leftarrow\texttt{cancelled}\);
          \item leave flight-level seat availability unchanged.
      \end{itemize}

      \item \texttt{postcondition:}
      \begin{itemize}
          \item the reservation is cancelled;
          \item refunds mirror the original payments and produce zero net payment;
          \item refunds settle within 5--7 business days.
      \end{itemize}

      \item \texttt{ambiguity:} the API does not enforce the flown-portion,
            cancellation-trigger, reason, or confirmation requirements; the
            agent must enforce these policy guards.
  \end{itemize}

  \medskip
  \texttt{\textbf{action\_spec: update\_reservation\_flights}} (relevant fields only)
  \begin{itemize}
      \setlength{\itemsep}{2pt}
      \item a cabin change is permitted if no segment has already been flown;
      \item the cabin must be uniform across all segments, sufficient seats must
            exist, and the user pays any fare increase;
      \item on success,
            \(\texttt{reservation.cabin}\leftarrow\texttt{requested cabin}\).
  \end{itemize}

  \end{artifactbox}

Given the guarded operators, we can directly identify policy interactions between flight modification and cancellation. For example, cancellation is blocked
  when an economy reservation is uninsured, was booked more than 24 hours ago, and is being cancelled for a change of plans. Upgrading the reservation to
  business class activates a separate cancellation trigger, making the same reservation eligible for cancellation:
\[
  \begin{aligned}
  &\underbrace{\text{economy}\wedge\text{uninsured}
  \wedge\text{old booking}}_{\text{cancellation blocked}}
  \xrightarrow{\text{upgrade cabin}}
  \underbrace{\text{business}}_{\text{trigger enabled}}
  \\[-1pt]
  &\underbrace{\text{business}}_{\text{trigger enabled}}
  \xrightarrow{\text{cancel}}
  \text{status}\leftarrow\text{cancelled}.
  \end{aligned}
  \]

\paragraph{Coverage-guided structural evolution.}
The following trace shows how each evolution step preserves the existing
  requirements while adding one new structural demand. 
  
  \begin{artifactbox}{Depth 0: Atomic cancellation lesson}
  The user asks to cancel an uninsured economy reservation booked more than
  24 hours ago because their plans changed. None of the four cancellation
  triggers holds, so the agent must refuse the cancellation.
  \end{artifactbox}

  \par\noindent\hfil$\Downarrow$\hfil\par
  \vspace{-2pt}

  \begin{artifactbox}{Depth 1: Add a logical chain}
  \textbf{Retained:} the original cancellation request is initially blocked.

  \textbf{Added:} The user asks any workarounds to get the flight canceled. The agent must discover an intermediate action that changes
  the governing policy condition. It first suggests to upgrade the reservation to business
  class, thereby activating the business-cabin cancellation trigger, and then
  completes the original cancellation:
  \[
  \texttt{upgrade\_to\_business}
  \;\longrightarrow\;
  \texttt{cancel\_reservation}.
  \]
  \end{artifactbox}

  \par\noindent\hfil$\Downarrow$\hfil\par
  \vspace{-2pt}

  \begin{artifactbox}{Depth 2: Add a parallel sub-task}
  \textbf{Retained:} the complete upgrade-cancel chain from Depth~1.

  \textbf{Added:} In addition to cancel the flight, the user separately asks to update baggage on another
  reservation. This request is independent of the cancellation, requiring the
  agent to manage two distinct entities and action types without mixing their
  state:
  \[
  \bigl(\texttt{upgrade}(A)\rightarrow\texttt{cancel}(A)\bigr)
  \quad+\quad
  \texttt{update\_baggage}(B).
  \]
  \end{artifactbox}

  \par\noindent\hfil$\Downarrow$\hfil\par
  \vspace{-2pt}

  \begin{artifactbox}{Depth 3: Extend the causal chain}
  \textbf{Retained:} the upgrade--cancel workflow and the independent baggage
  update.

  \textbf{Added:} After cancelling the original trip and updating baggages on another trip, the user requests
  replacement travel. The agent must search the environment and determine that
  the requested booking cannot be completed because the target flight lacks
  sufficient seats:
  \[
  \texttt{update\_baggage}(B)
  \quad+\quad
  \texttt{upgrade}(A)\rightarrow\texttt{cancel}(A)
  \rightarrow\texttt{replacement denied}.
  \]
  \end{artifactbox}

  \par\noindent\hfil$\Downarrow$\hfil\par
  \vspace{-2pt}

  \begin{artifactbox}{Depth 4: Expand cancellation into an exception-set loop}
  \textbf{Retained:} the cabin-upgrade dependency, baggage update, and failed
  replacement booking.

  \textbf{Added:} the single cancellation is generalized to
  ``cancel all my reservations except \(B\).'' The agent must enumerate the
  user's reservations, preserve the named exception, and apply the cancellation
  policy separately to every remaining reservation. Different iterations produce
  different outcomes: some reservations can be cancelled directly, one requires
  the upgrade--cancel chain, and another cannot be cancelled because a segment
  has already been flown.
  \end{artifactbox}

  \par\noindent\hfil$\Downarrow$\hfil\par
  \vspace{-2pt}

  \begin{artifactbox}{Final step: Add interaction variation}
  \textbf{Retained:} all structural actions, dependencies, exceptions, and
  outcomes from Depth~4.

  \textbf{Added:} the same requirements are surfaced less directly. The user
  buries an invalid compensation request among valid operations, accepts the
  cabin upgrade only if its cost is below a threshold, describes the baggage
  quantity relative to an unknown free allowance, and proceeds only when the
  resulting fee is at most \$100. The final task therefore requires policy
  resistance, state-dependent parameter resolution, conditional action, and
  boundary reasoning without changing its underlying structural lesson.
  \end{artifactbox}

\paragraph{Task initialization.}
Initialization grounds the evolved blueprint in executable entities, prices, availability, and user state. The first realization selected a gold member with five reservations and instantiated the following task.

\begin{artifactbox}{Round-0 initialized task}
\small
\textbf{User request.} ``Cancel all my reservations except \texttt{6DHUJY}.'' For reservation \texttt{55VQNU}, upgrade to business and cancel only if the upgrade costs less than \$300. After the cancellations, request compensation for the inconvenience and push back twice if it is refused. Then add two bags beyond the free allowance to \texttt{6DHUJY}, but only if the fee is at most \$100. Finally, request a one-way economy replacement flight from Denver to Chicago.

\medskip
\textbf{Grounded state.}
\begin{itemize}
    \setlength{\itemsep}{1pt}
    \item membership: gold;
    \item \texttt{55VQNU}: economy, uninsured, upgrade cost \$100;
    \item \texttt{U8XRWI}: business cabin and directly cancellable;
    \item \texttt{9FNHDR}: intended to contain a previously flown segment;
    \item \texttt{6DHUJY}: business cabin with four free bags; six total bags incur exactly \$100;
    \item intended replacement outcome: no acceptable economy seats.
\end{itemize}

\textbf{Intended write outcomes.}
\[
\begin{aligned}
&\texttt{upgrade(55VQNU)}
 \rightarrow \texttt{cancel(55VQNU)},\\
&\texttt{cancel(U8XRWI)},\\
&\texttt{refuse\_cancel(9FNHDR;\ flown\ portion)},\\
&\texttt{update\_baggage(6DHUJY;\ 6\ total,\ \$100)},\\
&\texttt{refuse\_replacement\_booking(insufficient\ seats)}.
\end{aligned}
\]
\end{artifactbox}

This realization preserves the evolved lesson: cancellation is initially blocked, a cabin update changes the governing policy branch, and the agent must also maintain an exception set and several independent constraints.

\paragraph{Reflection-guided solution search.}
The first rollout did not immediately solve the task. Rather than accepting it as a demonstration or discarding the lesson, the evaluator returned policy-grounded feedback for another simulation attempt.
\begin{artifactbox}{Simulation and reflection trace}
  \small
  \textbf{Initial rollout.}
  The first trajectory did not satisfy the evaluator, so its feedback was returned
  to the proposal model rather than accepting the demonstration.

  \medskip
  \textbf{Reflection retry 1: incomplete policy verification.}
  The revised agent still denied cancellation of reservation \texttt{9FNHDR} by
  inferring that an earlier flight date meant the segment had been flown, without
  checking its recorded status.

  \emph{Evaluator feedback:}
  ``Before denying cancellation for a flown portion, call
  \texttt{get\_flight\_status} and verify that the segment is
  \texttt{flying} or \texttt{landed}; do not infer its status from the date.''

  \medskip
  \textbf{Reflection retry 2: corrected behavior.}
  The agent incorporated the additional feedback and checked the flight status.
  It discovered that the segment was marked \texttt{cancelled}, activating the
  airline-cancellation trigger rather than the flown-portion prohibition. The
  agent then correctly completed the cancellation workflow, including the
  upgrade--cancel chain, named exception, compensation refusal, and baggage-price
  boundary.

  \medskip
  \textbf{Result.}
  The evaluator assigned the revised trajectory a score of \(1.0\), but identified
  the reference as incorrect: the initialized environment did not contain the
  flown segment assumed by the proposed solution.
  \end{artifactbox}

  The two retries illustrate that reflection need not repair every error at once:
  each evaluation narrows the remaining failure. Once the agent-side behavior is
  corrected, the residual mismatch supplies evidence for routing the example back
  to task initialization.

\paragraph{Verification and task-side routing.}
Execution verification compared the corrected rollout with the proposed reference state and found that the disagreement was caused by initialization, not by the agent.

\begin{artifactbox}{Round-0 verification result}
\small
\textbf{Route:} \texttt{initialization}

\medskip
\textbf{Diagnosis 1.} The task intended \texttt{9FNHDR} to contain a flown segment, but flight \texttt{HAT277} was initialized as \texttt{cancelled}, not \texttt{landed} or \texttt{flying}. The agent therefore correctly treated the airline-cancelled flight as a valid cancellation trigger.

\medskip
\textbf{Diagnosis 2.} The reference assumed that replacement travel could not be booked. However, the request did not rule out basic-economy or connecting alternatives, and the executable environment contained an alternative the user could legitimately accept.

\medskip
\textbf{Recovery instruction.} Re-initialize the task with a genuinely flown segment and make the replacement flight constraint explicit, or update the reference to include the available alternative.
\end{artifactbox}

\textsc{Gradus} returns it to initialization while keeping its structural lineage and interaction requirements fixed.

\paragraph{Recovery across initialization rounds.}
Table~\ref{tab:e2e-recovery} summarizes how simulation feedback is converted into task redesign.

\begin{table}[h]
\centering
\small
\setlength{\tabcolsep}{4pt}
\begin{tabular}{@{}p{0.12\linewidth}p{0.37\linewidth}p{0.43\linewidth}@{}}
\toprule
Round & Verification finding & Recovery \\
\midrule
0
& The supposed flown segment was actually airline-cancelled, and the replacement request admitted valid alternatives.
& Re-ground the same blueprint with a user having a truly landed segment, sold-out direct economy inventory, and reference actions consistent with the executable state. \\

1
& The flown-segment defect was repaired, but the user could still accept a connecting economy itinerary because only basic economy had been excluded.
& Clarify that the replacement request is for a \emph{direct} economy flight and that connecting and basic-economy alternatives must be declined. \\

2
& The predicted and reference database hashes match: \texttt{agent\_db\_match=true}.
& Accept the recovered task into the final task set. \\
\bottomrule
\end{tabular}
\caption{Task-side recovery preserves the evolved lesson while repairing grounding and ambiguity defects.}
\label{tab:e2e-recovery}
\end{table}

This trace illustrates the coupling between search and recovery. Structural evolution discovers a difficult policy interaction and expands it into a multi-constraint lesson; reflection repairs the demonstration; and verification uses residual disagreement to repair the task itself. The final example therefore retains the upgrade--cancel dependency, exception loop, policy resistance, and boundary reasoning that a one-shot filtering pipeline would have discarded.
\begin{artifactbox}{Final recovered task}
\small
The recovered task retains the original difficulty:

\begin{itemize}
    \setlength{\itemsep}{2pt}
    \item an uninsured economy reservation is upgraded to business because the \$273 cost is below the user's \$300 threshold, then cancelled;
    \item two existing business reservations are cancelled directly;
    \item a reservation with a genuinely landed segment must not be cancelled;
    \item one named reservation is excluded from the cancellation loop;
    \item the agent refuses compensation for user-initiated cancellations despite two misleading appeals;
    \item the kept reservation receives six bags: four free and two paid, placing the \$100 fee exactly at the acceptance boundary; and
    \item the direct replacement booking is denied because no direct economy seats remain, and the user explicitly declines connecting or basic-economy alternatives.
\end{itemize}

The final execution reaches exactly the proposed state (\texttt{predicted\_hash = gold\_hash}).
\end{artifactbox}

\subsection{Curriculum Transfer Across Student Sizes}
\label{app:scaling}

Tables~\ref{tab:scaling-tau} and~\ref{tab:scaling-car} compare dense 8B and 14B students with the 30B-A3B mixture-of-experts student, rather than conventional model-size scaling. The same curriculum transfers across all three architectures, with the MoE model showing its clearest advantage on CAR disambiguation, while gains on $\tau^2$-Bench are non-monotonic. Curricula proposed by Opus~4.5 generally outperform those from Qwen3-235B across students. Finally, outcome reward-based RL improves the MoE checkpoint on every split, with the largest gain on CAR hallucination.

\begin{table}[h]
\caption{Student architecture and capacity comparison on $\tau^2$-Bench.}
\label{tab:scaling-tau}
\begin{center}
\small
\begin{tabular}{llcccc}
\toprule
Model & Stage & Proposer & Airline & Retail & Telecom \\
\midrule

Qwen3-8B (Ours) & SFT & Qwen3-235B & \ms{0.71}{0.02} & \ms{0.74}{0.02} & \ms{0.64}{0.01} \\
Qwen3-14B (Ours) & SFT & Qwen3-235B & \ms{0.69}{0.04} & \ms{0.85}{0.02} & \ms{0.69}{0.03} \\
Qwen3-30B-A3B (Ours)& SFT & Qwen3-235B & \ms{0.69}{0.05} & \ms{0.85}{0.02} & \ms{0.75}{0.02} \\
Qwen3-8B (Ours)& SFT & Opus 4.5 & \ms{0.71}{0.05} & \ms{0.85}{0.02} & \ms{0.92}{0.01} \\
Qwen3-14B (Ours) & SFT & Opus 4.5 & \ms{0.75}{0.01} & \ms{0.85}{0.01} & \ms{0.85}{0.03} \\
Qwen3-30B-A3B (Ours)& SFT & Opus 4.5 & \ms{0.75}{0.05} & \ms{0.86}{0.03} & \ms{0.90}{0.02} \\
\midrule
Qwen3-30B-A3B  (Ours) & SFT+RL & Opus 4.5 & \ms{0.80}{0.03} & \ms{0.89}{0.02} & \ms{0.97}{0.01} \\

\bottomrule
\end{tabular}
\end{center}
\end{table}

\begin{table}[h]
\caption{Student architecture and capacity comparison on Car-Bench.}
\label{tab:scaling-car}
\begin{center}
\small
\begin{tabular}{llcccc}
\toprule
Model & Stage & Proposer & Base & Disamb. & Halluc. \\
\midrule

Qwen3-8B (Ours) & SFT & Qwen3-235B & \ms{0.69}{0.04} & \ms{0.45}{0.04} & \ms{0.51}{0.03} \\
Qwen3-14B (Ours) & SFT & Qwen3-235B & \ms{0.68}{0.03} & \ms{0.50}{0.06} & \ms{0.55}{0.04} \\
Qwen3-30B-A3B  (Ours) & SFT & Qwen3-235B & \ms{0.72}{0.04} & \ms{0.60}{0.05} & \ms{0.55}{0.04} \\

Qwen3-8B (Ours) & SFT & Opus 4.5 & \ms{0.76}{0.02} & \ms{0.58}{0.06} & \ms{0.74}{0.04} \\
Qwen3-14B (Ours) & SFT & Opus 4.5 & \ms{0.80}{0.02} & \ms{0.60}{0.03} & \ms{0.74}{0.02} \\
Qwen3-30B-A3B  (Ours) & SFT & Opus 4.5 & \ms{0.80}{0.04} & \ms{0.74}{0.02} & \ms{0.74}{0.02} \\
\midrule
Qwen3-30B-A3B (Ours) & SFT+RL & Opus 4.5 & \ms{0.83}{0.03} & \ms{0.75}{0.05} & \ms{0.91}{0.02} \\

\bottomrule
\end{tabular}
\end{center}
\end{table}

\subsection{Experiment Results (Detailed)}
\label{sec:main_result_details}

Tables~\ref{tab:tau2} and~\ref{tab:car} supplement the main results with five-run standard deviations and $\mathrm{pass}^{3}$, which credits a task only when all three independent runs succeed~\citep{barres2025tau2benchevaluatingconversationalagents,kirmayr-etal-2026-car}. This metric penalizes intermittent behavior. The overall ranking is preserved: on $\tau^2$, RL decreases from $0.89$ mean success to $0.78$ $\mathrm{pass}^{3}$, turning its near-parity in mean success with Opus~4.5 into a four-point reliability gap. On CAR, however, its frontier lead remains ($0.71$ versus $0.64$). The remaining instability is concentrated on CAR disambiguation across frontier models and on airline for GRADUS.

\begin{table}[h]
\caption{Detailed task success on $\tau^2$-Bench. Cells report mean $\pm$ standard deviation
over five runs. Avg averages the three domains. $\mathrm{pass}^k$ credits a task only when all
$k$ independent runs succeed;we report $k=3$. Proposer names the curriculum-construction model
where applicable. GRADUS rows are shaded; \textbf{bold} marks the best value outside the
\emph{Frontier models} block.}
\label{tab:tau2}
\begin{center}
\small
\setlength{\tabcolsep}{4pt}
\begin{tabular}{ll ccccc}
\toprule
Model & Proposer & Airline & Retail & Telecom & Avg & Avg.\ $\mathrm{pass}^3$ \\
\midrule
\multicolumn{7}{l}{\emph{Frontier models}} \\
Opus 4.5         & -- & \ms{0.88}{0.01} & \ms{0.89}{0.01} & \ms{0.92}{0.01} & $0.90$ & $0.82$ \\
GPT-5.2          & -- & \ms{0.82}{0.05} & \ms{0.88}{0.02} & \ms{0.89}{0.01} & $0.86$ & $0.76$ \\
Kimi-K2-Thinking & -- & \ms{0.84}{0.03} & \ms{0.88}{0.01} & \ms{0.79}{0.03} & $0.84$ & $0.73$ \\
DeepSeek-V3.2    & -- & \ms{0.78}{0.03} & \ms{0.91}{0.02} & \ms{0.76}{0.04} & $0.82$ & $0.66$ \\
\midrule
\multicolumn{7}{l}{\emph{Open-weight backbones}} \\
Qwen3-8B   & -- & \ms{0.38}{0.04} & \ms{0.41}{0.02} & \ms{0.40}{0.04} & $0.40$ & $0.22$ \\
Qwen3-14B  & -- & \ms{0.42}{0.04} & \ms{0.60}{0.03} & \ms{0.42}{0.02} & $0.48$ & $0.31$ \\
Qwen3-30B  & -- & \ms{0.63}{0.04} & \ms{0.77}{0.01} & \ms{0.38}{0.02} & $0.59$ & $0.42$ \\
Qwen3-235B & -- & \ms{0.61}{0.03} & \ms{0.80}{0.02} & \ms{0.50}{0.03} & $0.64$ & $0.44$ \\
\midrule
\multicolumn{7}{l}{\emph{Generalist agent training (released checkpoints)}} \\
EnvScaler (8B)        & -- & \ms{0.36}{0.04} & \ms{0.57}{0.03} & \ms{0.33}{0.02} & $0.42$ & $0.23$ \\
Klear-AgentForge (8B) & -- & \ms{0.39}{0.06} & \ms{0.70}{0.04} & \ms{0.26}{0.01} & $0.45$ & $0.27$ \\
AgentSkiller (8B)      & -- & \ms{0.34}{0.06} & \ms{0.64}{0.03} & \ms{0.65}{0.02} & $0.54$ & $0.30$ \\
AgenticQwen (8B)      & -- & \ms{0.44}{0.02} & \ms{0.64}{0.05} & \ms{0.45}{0.01} & $0.51$ & $0.31$ \\
\midrule
\multicolumn{7}{l}{\emph{Environment-targeted synthesis (matched student \& budget)}} \\
Simia (8B) & GPT-5/o4-mini & \ms{0.36}{0.04} & \ms{0.60}{0.03} & -- & -- & -- \\
CoVe (8B)  & Qwen3-235B    & \ms{0.31}{0.02} & \ms{0.66}{0.04} & -- & -- & -- \\
WRIT (8B)  & Qwen3.6-Plus  & \ms{0.58}{0.03} & \ms{0.78}{0.03} & -- & -- & -- \\
\midrule
\multicolumn{7}{l}{\emph{Environment-targeted synthesis (matched proposer, student \& budget)}} \\
Model-Prior Synthesis (8B)      & Opus-4.5 & \ms{0.46}{0.04} & \ms{0.53}{0.03} & \ms{0.57}{0.02} & $0.52$ & $0.33$ \\
Graph-Based Synthesis (8B)      & Opus-4.5 & \ms{0.53}{0.03} & \ms{0.59}{0.03} & \ms{0.62}{0.03} & $0.58$ & $0.41$ \\
Trajectory-First Synthesis (8B) & Opus-4.5 & \ms{0.59}{0.03} & \ms{0.70}{0.03} & \ms{0.71}{0.02} & $0.67$ & $0.43$ \\
\midrule
\multicolumn{7}{l}{\emph{GRADUS (ours)}} \\
\rowcolor{blue!8} Qwen3-8B & Qwen3-235B & \ms{0.71}{0.02} & \ms{0.74}{0.02} & \ms{0.64}{0.01} & $0.70$ & $0.51$ \\
\addlinespace[1pt]
\rowcolor{blue!8} Qwen3-8B & Opus-4.5 & \ms{0.71}{0.05} & \ms{0.85}{0.02} & \ms{0.92}{0.01} & $0.83$ & $0.68$ \\
\addlinespace[1pt]
\rowcolor{blue!8} Qwen3-30B-A3B & Qwen3-235B & \ms{0.69}{0.05} & \ms{0.85}{0.02} & \ms{0.75}{0.02} & $0.76$ & $0.61$ \\
\addlinespace[1pt]
\rowcolor{blue!8} Qwen3-30B-A3B & Opus-4.5 & \ms{0.75}{0.05} & \ms{0.86}{0.03} & \ms{0.90}{0.02} & $0.84$ & $0.71$ \\
\addlinespace[1pt]
\rowcolor{blue!8} Qwen3-30B-A3B +RL & Opus-4.5 & \ms{\mathbf{0.80}}{0.03} & \ms{\mathbf{0.89}}{0.02} & \ms{\mathbf{0.97}}{0.01} & $\mathbf{0.89}$ & $\mathbf{0.78}$ \\
\bottomrule
\end{tabular}
\end{center}
\end{table}

\begin{table}[h]
\caption{Detailed task success on CAR-Bench. Cells report mean $\pm$ standard deviation over
five runs. Avg averages the three splits. $\mathrm{pass}^k$ credits a task only when all $k$
independent runs succeed; we report $k=3$. External curriculum baselines (Simia, CoVe, WRIT)
do not report CAR-Bench and are omitted. Proposer names the curriculum-construction model
where applicable. GRADUS rows are shaded; \textbf{bold} marks the best value outside the
\emph{Frontier models} block.}
\label{tab:car}
\begin{center}
\small
\setlength{\tabcolsep}{4pt}
\begin{tabular}{ll ccccc}
\toprule
Model & Proposer & Base & Disamb & Halluc & Avg & Avg.\ $\mathrm{pass}^3$ \\
\midrule
\multicolumn{7}{l}{\emph{Frontier models}} \\
Opus 4.5         & -- & \ms{0.85}{0.02} & \ms{0.68}{0.06} & \ms{0.64}{0.03} & $0.72$ & $0.61$ \\
GPT-5.2          & -- & \ms{0.79}{0.02} & \ms{0.52}{0.08} & \ms{0.88}{0.01} & $0.73$ & $0.64$ \\
Kimi-K2-Thinking & -- & \ms{0.56}{0.03} & \ms{0.42}{0.05} & \ms{0.67}{0.03} & $0.55$ & $0.43$ \\
DeepSeek-V3.2    & -- & \ms{0.64}{0.02} & \ms{0.48}{0.07} & \ms{0.57}{0.03} & $0.56$ & $0.41$ \\
\midrule
\multicolumn{7}{l}{\emph{Open-weight backbones}} \\
Qwen3-8B   & -- & \ms{0.40}{0.04} & \ms{0.32}{0.05} & \ms{0.40}{0.04} & $0.37$ & $0.22$ \\
Qwen3-14B  & -- & \ms{0.49}{0.05} & \ms{0.35}{0.06} & \ms{0.42}{0.06} & $0.42$ & $0.29$ \\
Qwen3-30B  & -- & \ms{0.52}{0.06} & \ms{0.43}{0.02} & \ms{0.45}{0.04} & $0.47$ & $0.30$ \\
Qwen3-235B & -- & \ms{0.64}{0.05} & \ms{0.46}{0.06} & \ms{0.46}{0.05} & $0.52$ & $0.39$ \\
\midrule
\multicolumn{7}{l}{\emph{Generalist agent training (released checkpoints)}} \\
EnvScaler (8B)        & -- & \ms{0.48}{0.05} & \ms{0.34}{0.05} & \ms{0.34}{0.02} & $0.39$ & $0.27$ \\
Klear-AgentForge (8B) & -- & \ms{0.46}{0.03} & \ms{0.29}{0.06} & \ms{0.36}{0.01} & $0.37$ & $0.24$ \\
AgentSkiller (8B)      & -- & \ms{0.49}{0.03} & \ms{0.34}{0.07} & \ms{0.33}{0.05} & $0.39$ & $0.25$ \\
AgenticQwen (8B)      & -- & \ms{0.55}{0.05} & \ms{0.32}{0.04} & \ms{0.56}{0.03} & $0.48$ & $0.32$ \\
\midrule
\multicolumn{7}{l}{\emph{Environment-targeted synthesis (matched proposer, student \& budget)}} \\
Model-Prior Synthesis (8B)      & Opus-4.5 & \ms{0.40}{0.03} & \ms{0.44}{0.05} & \ms{0.31}{0.04} & $0.38$ & $0.24$ \\
Graph-Based Synthesis (8B)      & Opus-4.5 & \ms{0.49}{0.04} & \ms{0.36}{0.04} & \ms{0.46}{0.03} & $0.44$ & $0.31$ \\
Trajectory-First Synthesis (8B) & Opus-4.5 & \ms{0.56}{0.03} & \ms{0.48}{0.04} & \ms{0.52}{0.03} & $0.52$ & $0.38$ \\
\midrule
\multicolumn{7}{l}{\emph{GRADUS (ours)}} \\
\rowcolor{blue!8} Qwen3-8B & Qwen3-235B & \ms{0.69}{0.04} & \ms{0.45}{0.04} & \ms{0.51}{0.03} & $0.55$ & $0.42$ \\
\addlinespace[1pt]
\rowcolor{blue!8} Qwen3-8B & Opus-4.5 & \ms{0.76}{0.02} & \ms{0.58}{0.06} & \ms{0.74}{0.04} & $0.69$ & $0.55$ \\
\addlinespace[1pt]
\rowcolor{blue!8} Qwen3-30B-A3B & Qwen3-235B & \ms{0.72}{0.04} & \ms{0.60}{0.05} & \ms{0.55}{0.04} & $0.62$ & $0.51$ \\
\addlinespace[1pt]
\rowcolor{blue!8} Qwen3-30B-A3B & Opus-4.5 & \ms{0.80}{0.04} & \ms{0.74}{0.02} & \ms{0.74}{0.02} & $0.76$ & $0.64$ \\
\addlinespace[1pt]
\rowcolor{blue!8} Qwen3-30B-A3B +RL & Opus-4.5 & \ms{\mathbf{0.83}}{0.03} & \ms{\mathbf{0.75}}{0.05} & \ms{\mathbf{0.91}}{0.02} & $\mathbf{0.83}$ & $\mathbf{0.71}$ \\
\bottomrule
\end{tabular}
\end{center}
\end{table}

\subsection{Evolved Frontier Exams: Construction and Human Audit}
\label{app:annotation}

\paragraph{Frontier Exam Construction.}
Starting from the original benchmark tasks, Opus~4.5 performs five depths of structural and interaction evolution. Each blueprint passes through the same initialization, simulation, and verification loop used for curriculum construction. Among candidates judged valid, we form the set of exam candidates whose reference outcome is matched in at most five of the eight rollouts. Table~\ref{tab:audit-funnel} summarizes the  workflow.

\begin{table}[h]
\centering
\small
\setlength{\tabcolsep}{5pt}
\begin{tabular}{@{}lrrrrrr@{}}
\toprule
Domain & Evolved & Executable & Valid & $\leq 5/8$  & Both correct \\
\midrule
Airline ($\tau^2$) & 212 & 84  & 82    & 32 & 25 \\
Retail ($\tau^2$)  & 399 & 225 & 222   & 26 & 26 \\
CAR-Bench & 512 & 472 & 324  & 79 & 70 \\
\bottomrule
\end{tabular}
\caption{Construction and human-audit counts. }
\label{tab:audit-funnel}
\end{table}

\paragraph{Annotation protocol.}
Two annotators independently assessed all 137 frozen candidates: 32 airline, 26 retail, and 79 CAR tasks. Before annotating each domain tasks, annotators read its policy and tool schema to understand the domain regulations and expected agent behaviors. For every task, they were given the task, initial states, and oracle action sequence (reference) and answered the following questions without conferring.

\medskip
\noindent
\fcolorbox{black!25}{gray!4}{
\begin{minipage}{0.94\linewidth}
\small
\newcommand{\optgap}{\hspace{2em}}
\textbf{Q1: Correctness.} Is the oracle outcome what a competent, policy-compliant assistant should produce?\\
A. Yes\optgap B. No\optgap C. Cannot tell \\[2pt]
\textbf{Q2: Realism.} Would a real user plausibly make this request in this situation and in this way?\\
A. Natural\optgap B. Plausible but unusual\optgap C. Contrived \\[2pt]
\textbf{Q3: Oracle strictness.} Could a policy-compliant assistant legitimately use different steps, parameters, or ordering while reaching the same outcome?\\
A. No\optgap B. Partly\optgap C. Yes, materially over-specified
\end{minipage}}
\medskip

\paragraph{Annotation results.}
Table~\ref{tab:annotation-labels} reports pooled label distributions, averaging the two annotators. Correctness is high across all domains: $87.5\%$ for airline, $100\%$ for retail, and $90.5\%$ for CAR. Both annotators marked 25/32, 26/26, and 70/79 tasks correct, respectively; only these jointly accepted tasks are used for the reported frontier-exam results. CAR requests are predominantly natural and have non-strict oracles. The evolved $\tau^2$ tasks are more often judged unusual or partly strict, reflecting greater subjectivity about scenario naturalness and interchangeable execution paths.

\begin{table}[h]
\centering
\small
\setlength{\tabcolsep}{6pt}
\begin{tabular}{@{}lccc@{}}
\toprule
Domain &
Correctness (\textsc{Yes/No/CT}) &
Realism (\textsc{Nat./Unusual/Contr.}) &
Strictness (\textsc{No/Partly/Yes}) \\
\midrule
Airline & 87.5 / 10.9 / 1.6 & 26.6 / 53.1 / 20.3 & 18.8 / 70.3 / 10.9 \\
Retail  & 100.0 / 0.0 / 0.0 & 46.2 / 32.7 / 21.2 & 53.8 / 38.5 / 7.7 \\
CAR     & 90.5 / 8.2 / 1.3  & 88.6 / 11.4 / 0.0 & 81.0 / 13.3 / 5.7 \\
\bottomrule
\end{tabular}
\caption{Pooled annotation distributions (\%). }
\label{tab:annotation-labels}
\end{table}

\paragraph{Inter-annotator agreement.}
  Raw agreement and Gwet's AC1 support the same conclusions
  (Table~\ref{tab:annotation-agreement}). Correctness agreement is substantial
  for airline (81.2\%, AC1 $=0.789$) and near-perfect for retail and CAR
  (96.2--100\%, AC1 $=0.958$-$1.000$). Agreement on oracle strictness is also
  strong (AC1 $=0.728$, $0.821$, and $0.986$), indicating that annotators
  generally agree on whether an oracle is materially over-specified. Realism is
  more subjective for airline and retail (AC1 $=0.399$ and $0.596$), but
  near-perfect for CAR (AC1 $=1.000$); the wider confidence intervals on
  $\tau^2$ reflect this uncertainty and the smaller audit sets. For realism and
  strictness, AC1 is computed over binary acceptable-versus-problematic
  judgments, while Table~\ref{tab:annotation-labels} retains the full label
  distributions~\citep{gwet2008computing}. Overall, the audit provides strong
  evidence for semantic correctness and non-strictness, with exception of disagreement concentrated in
  subjective judgments of naturalness on Airline and Retail domain.

  \begin{table}[h]
  \centering
  \small
  \setlength{\tabcolsep}{6pt}
  \begin{tabular}{@{}llcc@{}}
  \toprule
  Domain & Dimension & Raw agreement & Gwet's AC1 [95\% CI] \\
  \midrule
  Airline & Correctness       & 81.2\%  & 0.789 [0.586, 0.933] \\
          & Realism           & 59.4\%  & 0.399 [0.056, 0.680] \\
          & Oracle strictness & 78.1\%  & 0.728 [0.461, 0.897] \\
  \midrule
  Retail  & Correctness       & 100.0\% & 1.000 [1.000, 1.000] \\
          & Realism           & 73.1\%  & 0.596 [0.239, 0.860] \\
          & Oracle strictness & 84.6\%  & 0.821 [0.584, 0.960] \\
  \midrule
  CAR     & Correctness       & 96.2\%  & 0.958 [0.900, 1.000] \\
          & Realism           & 100.0\% & 1.000 [1.000, 1.000] \\
          & Oracle strictness & 98.7\%  & 0.986 [0.954, 1.000] \\
  \bottomrule
  \end{tabular}
  \caption{Inter-annotator agreement. }
  \label{tab:annotation-agreement}
  \end{table}

\end{document}